\pdfoutput=1
\documentclass[11pt]{article}

\usepackage[final]{acl}

\usepackage{times}
\usepackage{latexsym}

\usepackage[T1]{fontenc}
\usepackage[utf8]{inputenc}

\usepackage{microtype}

\usepackage{inconsolata}
\usepackage{todonotes}
\usepackage{booktabs}
\usepackage{multirow}
\usepackage{enumitem}
\usepackage{makecell}
\usepackage{pifont}
\usepackage{longtable}
\usepackage{graphicx}
\usepackage{devanagari}
\usepackage{CJKutf8}
\usepackage{tablefootnote}
\title{SPEED++: A Multilingual Event Extraction Framework for \\ Epidemic Prediction and Preparedness}

\author{Tanmay Parekh \ \ \ \ \
Jeffrey Kwan \ \ \ \ \
Jiarui Yu \ \ \ \ \
Sparsh Johri \ \ \ \ \ 
Hyosang Ahn \\
{\bf Sreya Muppalla \ \ \ \ \
Kai-Wei Chang \ \ \ \ \
Wei Wang \ \ \ \ \
Nanyun Peng} \\
Computer Science Department, University of California, Los Angeles \\
\texttt{\{tparekh, weiwang, violetpeng, kwchang\}@cs.ucla.edu} \\
  }

\begin{document}
\maketitle

\newcommand{\mypar}[1]{\vspace{0.35em}\noindent\textbf{#1}}
\newcommand{\SideNote}[2]{\todo[color=#1,size=\small]{#2}} 

\newcommand{\tanmay}[1]{\SideNote{orange!40}{#1 --tanmay}}
\newcommand{\kaiwei}[1]{\SideNote{brown!40}{#1 --kai-wei}}
\newcommand{\violet}[1]{\SideNote{purple!40}{#1 --violet}}
\newcommand{\wei}[1]{\SideNote{blue!40}{#1 --wei}}
\newcommand{\jeffrey}[1]{\SideNote{pink!40}{#1 --jeffrey}}
\newcommand{\hyosang}[1]{\SideNote{cyan!40}{#1 --hyosang}}
\newcommand{\sparsh}[1]{\SideNote{yellow!40}{#1 --sparsh}}

\newcommand{\dataName}[0]{SPEED++}

\newcommand{\cmark}{\ding{51}}%
\newcommand{\xmark}{\ding{55}}%
\newcommand{\tildemark}{\textbf{$\sim$}}%
\newcommand\splus{\ding{58}}

\newcommand{\redtext}[1]{\textcolor{red}{#1}}
\newcommand{\bluetext}[1]{\textcolor{blue}{#1}}
\newcommand{\browntext}[1]{\textcolor{brown}{#1}}

\newcommand{\hi}[1]{\dn{#1}}
\newcommand{\ja}[1]{\begin{CJK}{UTF8}{min}#1\end{CJK}}
\newcommand{\zh}[1]{\begin{CJK}{UTF8}{gbsn}#1\end{CJK}}

\newcommand\blfootnote[1]{%
  \begingroup
  \renewcommand\thefootnote{}\footnote{#1}%
  \addtocounter{footnote}{-1}%
  \endgroup
}

\begin{abstract}

Social media is often the first place where communities discuss the latest societal trends.
Prior works have utilized this platform to extract epidemic-related information (e.g. infections, preventive measures) to provide early warnings for epidemic prediction.
However, these works only focused on English posts, while epidemics can occur anywhere in the world, and early discussions are often in the local, non-English languages.
In this work, we introduce the first multilingual Event Extraction (EE) framework \dataName{} for extracting epidemic event information for a wide range of diseases and languages.
To this end, we extend a previous epidemic ontology with 20 argument roles; and curate our multilingual EE dataset \dataName{} comprising 5.1K tweets in four languages for four diseases.
Annotating data in every language is infeasible; thus we develop zero-shot cross-lingual cross-disease models (i.e., training only on English COVID data) utilizing multilingual pre-training and show their efficacy in extracting epidemic-related events for 65 diverse languages across different diseases.
Experiments demonstrate that our framework can provide epidemic warnings for COVID-19 in its earliest stages in Dec 2019 (3 weeks before global discussions) from Chinese Weibo posts without any training in Chinese.
Furthermore, we exploit our framework's argument extraction capabilities to aggregate community epidemic discussions like symptoms and cure measures, aiding misinformation detection and public attention monitoring.
Overall, we lay a strong foundation for multilingual epidemic preparedness.
% This paper has been accepted at EMNLP 2024.
% ~\footnote{Data will be released upon acceptance.}

\end{abstract}

\section{Introduction}

Timely epidemic-related information is vital for policymakers to issue warnings and implement control measures \cite{DBLP:journals/bioinformatics/CollierDKGCTNDKTST08}.
Social media being timely, publicly accessible, widely used, and high in volume \cite{heymann2001hot, lamb-etal-2013-separating, DBLP:journals/jbi/LybargerOTY21} acts as a crucial information source.
Previous works \cite{speed, zong-etal-2022-extracting} have explored utilizing Event Extraction (EE) \cite{sundheim-1992-overview, doddington-etal-2004-automatic} to extract epidemic events from social media posts for epidemic prediction.
However, these works have focused only on English; while epidemics can originate anywhere worldwide and be discussed in various regional languages.
% Since epidemics can arise anywhere, these social media posts can be in any language.
% Therefore, an automated system that monitors social media to extract epidemic-related information in any language is essential for effective global epidemic preparedness.

% To the end, Event Extraction (EE) \cite{sundheim-1992-overview, doddington-etal-2004-automatic} provides a robust method for extracting structured, event-specific information from natural language text.
% Recently, SPEED \cite{speed} developed an Event Detection (subtask of EE) framework for epidemic prediction by providing early warnings for emerging diseases. 
% However, SPEED is limited to English and is a partial EE framework as it cannot extract detailed event-specific information.

% \begin{figure}[t]
%     \centering
%     \includegraphics[width=0.48\textwidth]{figures/COVID_zh_preds_flattened.pdf}
%     \caption{Illustration of multilingual epidemic prediction in Chinese for COVID-19 pandemic in Dec-Jan 2020. Arrows indicate how our \dataName{} framework could provide warnings three weeks before its widespread discussion globally.}
%     \label{fig:zh-timeseries}
% \end{figure}

\begin{figure}[t]
    \centering
    \includegraphics[width=0.48\textwidth]{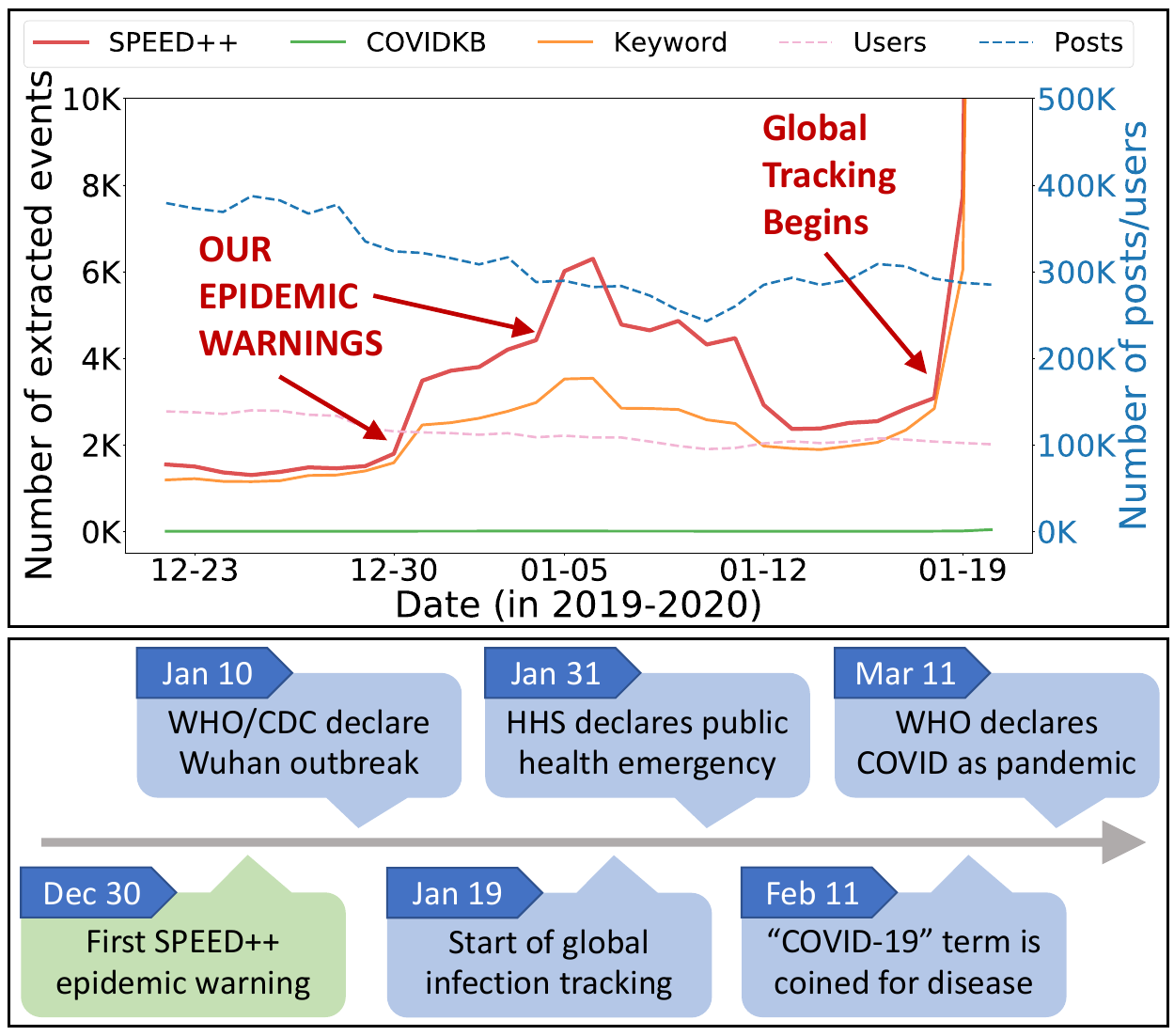}
    \caption{Zero-shot multilingual epidemic prediction in Chinese for COVID-19 pandemic.
    (Top) Number of epidemic events extracted in Dec-Jan 2020. Arrows indicate \dataName{} epidemic warnings. (Bottom) \dataName{} warning with respect to the general timeline of major moments of the COVID-19 pandemic.}
    \label{fig:zh-timeseries}
\end{figure}

% However their work has two major drawbacks which limits its practical utility.
% % Recently, SPEED built the first ED system and showed capabilities to provide early epidemic warnings - but their work has two major drawbacks which limit its utility. 
% First, SPEED focused solely on English - but diseases can originate anywhere in the world and people can talk about it in any language.
% Thus, there's a need for multilingual abilities in the extraction systems that can predict epidemics in any language.
% Secondly, SPEED is a ED system that only extracts the presence of events without any fetching any epidemic-specific information.
% Such epidemic-specific event information is vital for understanding different aspects of the predicted epidemic, for example, the associated disease, time and place of different events, symptoms people are facing, preventive measures, etc.
% This underlines the need for extracting epidemic-specific information to provide better signals to tackle the epidemic.
% \tanmay{Any line on other methods?}

In our work, we introduce \textbf{\dataName{}} (\textbf{S}ocial \textbf{P}latform based \textbf{E}pidemic \textbf{E}vent \textbf{D}etection \textbf{+} Arguments \textbf{+} Multilinguality), the first multilingual EE framework designed for epidemic preparedness.
% that can extract fine-grained epidemic event information from social media posts (specifically Twitter) for any disease and any language worldwide.
We advance English-based SPEED \cite{speed} multilingually by developing and benchmarking zero-shot cross-lingual models capable of extracting epidemic information across many languages.
While SPEED primarily identifies basic epidemic events, we develop enhanced models capable of extracting detailed event-specific information (e.g., \textit{symptoms}, \textit{control measures}) by incorporating Event Argument Extraction (EAE).
% Our framework advances upon SPEED \cite{speed} by enhancing model capabilities to extract event-specific information by integrating Event Argument Extraction (EAE) and the development of zero-shot cross-lingual models to extract information from any language.
% First, we enrich the ED only SPEED dataset with EAE annotations and this enables to fetch epidemic-event specific information.
To integrate EAE, we enrich the SPEED ontology with event-specific roles relevant to social media to create a rich EE ontology comprising 7 event types (e.g. \textit{infect}, \textit{cure}, \textit{prevent}, etc.) and 20 argument roles (e.g. \textit{disease}, \textit{symptoms}, \textit{time}, \textit{means}, etc.).
Apart from English, we also annotated three other languages - Spanish, Hindi, and Japanese - to benchmark multilingual EE models.
Leveraging the enriched ontology and expert annotations, we develop our \dataName{} dataset comprising 5.1K tweets and 4.6K event mentions across four different diseases (COVID-19, Monkeypox, Zika, and Dengue) in four languages.
% \tanmay{Kai-Wei: Remove mentions of "extending SPEED"}

Using \dataName{}, we develop our zero-shot cross-lingual models by empowering the state-of-the-art EE models like TagPrime \cite{hsu-etal-2023-tagprime} with multilingual pre-training and augmented training using pseudo-generated multilingual data from CLaP \cite{clap}.
% Leveraging our \dataName{} dataset and state-of-the-art EE models TagPrime \cite{hsu-etal-2023-tagprime} and CLaP \cite{clap}, we develop zero-shot cross-lingual cross-disease EE models.
These models are trained realistically on limited English COVID-specific data.
Benchmarking on \dataName{} reveals how our trained models outperform various baselines by an average of 15-16\% F1 points for unseen diseases across four different languages.
% Experiments demonstrate that our trained models outperform various baselines by an average 15-16\% F1 points.
% At the same time, our models are lightweight with XX parameters and capable of analyzing XX tweets within YY hours on a ZZ machine, making them highly efficient for practical deployment.

To demonstrate the utility of our multilingual EE \dataName{} framework, we apply it to two epidemic-related applications.
First, we utilize the framework's multilingual capabilities for epidemic prediction by aggregating epidemic events across different languages.
By incorporating tweet locations, we construct a global epidemic severity meter capable of providing epidemic warnings in 65 languages spanning 117 countries.
Applying our framework for COVID-19 to Chinese Weibo posts, we successfully detected early epidemic warnings by Dec 30, 2019 (Figure~\ref{fig:zh-timeseries}) - three weeks before the global infection tracking even began.
This multilingual epidemic prediction capability can significantly enhance our global preparedness for future epidemics.

As another application, we repurpose our framework as an information aggregation system for community discussions about epidemics such as \textit{symptoms, cure measures}, etc.
Leveraging the EAE capability of our framework, we meticulously extract these event-specific details from millions of tweets across diseases and languages.
Similar arguments are then agglomeratively clustered to generate an aggregated ranked bulletin.
We demonstrate that this bulletin can aid misinformation detection (e.g., \textit{cow urine} as a cure for \textit{COVID-19}) and public attention shift monitoring (e.g., \textit{rashes} as symptoms for \textit{Monkeypox}).
Such an automated disease-agnostic multilingual aggregation system can significantly alleviate human effort while providing insights into public epidemic opinions.

In conclusion, our work presents a three-fold contribution.
First, we create the first multilingual Event Extraction dataset for epidemic prediction SPEED++ encompassing four diseases and four languages.
Second, leveraging SPEED++, we develop models proficient in extracting epidemic-related data across a wide set of diseases and languages.
Lastly, we demonstrate the robust utility of our framework through two epidemic-centric applications, facilitating multilingual epidemic prediction and the aggregation of epidemic information.
% We release our data at \url{https://github.com/PlusLabNLP/SPEED-plus-plus}.

\section{Background}

\begin{figure}
    \centering
    \includegraphics[width=0.48\textwidth]{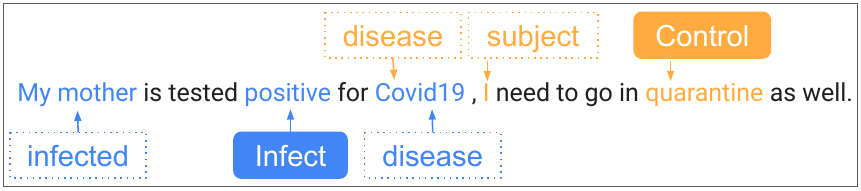}
    \caption{Illustration of Event Extraction for epidemic-related events \textit{Infect} and \textit{Control}. Corresponding arguments and their roles are marked in dotted boxes - that are absent in the SPEED \cite{speed} dataset.}
    \label{fig:ee-example}
\end{figure}

Epidemic prediction is a classic epidemiological task that provides early warnings for future epidemics of any infectious disease \cite{signorini2011use}.
% In our work, we focus the information source on social media posts, specifically Twitter.
Previous works \cite{DBLP:journals/artmed/LejeuneBDL15, DBLP:journals/jbi/LybargerOTY21} have utilized keyword-based and simple classification-based methods for extracting epidemic mentions (detailed in \S~\ref{sec:related-works}).
SPEED \cite{speed} was the first to explore Event Extraction (EE) for extracting epidemic-based events in English.
In our work, we utilize Event Extraction but focus multilingually on a broader range of languages.
The extracted events are aggregated over time and abnormal influxes are reported as early epidemic warnings.
To our best knowledge, we are the first to develop a multilingual Event Extraction framework for epidemic prediction.

\paragraph{Task Definition}
We adhere to the ACE 2005 guidelines \cite{doddington-etal-2004-automatic} to define an \textbf{event} as an occurrence or change of state associated with a specific \textbf{event type}.
An \textbf{event mention} is the sentence that describes the event, and it includes an \textbf{event trigger}, the word or phrase that most clearly indicates the event.
Event Extraction comprises two subtasks: Event Detection and Event Argument Extraction.
\textbf{Event Detection (ED)} involves identifying these event triggers in sentences and classifying them into predefined event types, while \textbf{Event Argument Extraction (EAE)} extracts arguments and assigns them event-specific roles.
Figure~\ref{fig:ee-example} shows an illustration for two event mentions for the events \textit{infect} and \textit{control}.

\section{Dataset Creation}

\begin{figure*}
    \centering
    \includegraphics[width=1\linewidth]{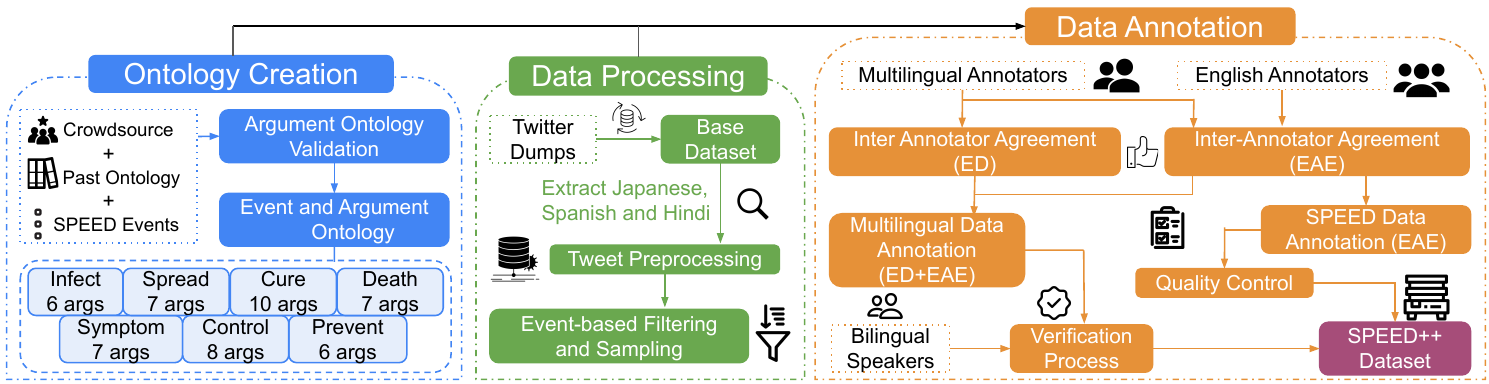}
    \caption{Overview of the data creation process. Majorly, we expand the ontology with argument roles, preprocess and filter the multilingual data, and annotate them using bilingual experts to create \dataName.}
    \label{fig:data-creation-process}
\end{figure*}

We focus on social media, specifically Twitter as our main document source for studying four diseases - COVID-19, Monkeypox, Zika, and Dengue.
SPEED \cite{speed} focused only on Event Detection (ED) for English.
Since ED identifies events but does not provide any epidemic-related information, we improve SPEED by additionally incorporating Event Argument Extraction (EAE) to develop a complete EE dataset \dataName.
Furthermore, we extend to three other
% \jeffrey{what does socially used mean}
languages used on social media - Spanish, Hindi, and Japanese - to enhance the multilingual capability of our framework.
We detail the data creation process below while Figure~\ref{fig:data-creation-process} provides a high-level overview.
% \tanmay{@jiarui Need a diagram similar to SPEED here (but a bit different as it may look too similar)}

% Mention the main additions - EAE and multilingual ED+EAE. Discuss how we selected languages for multilingual benchmarking - based on data availability and annotator availability. Also, discuss the major diseases as part of the data.

\subsection{Ontology Creation}
\label{sec:ontology-creation}

Event ontologies comprises event types and corresponding event-specific roles.
For our ontology, we derive the event types from SPEED and augment them with event-specific roles in our work.
We follow ACE guidelines \cite{doddington-etal-2004-automatic} for role definitions while also including a few non-entity roles based on GENEVA \cite{parekh-etal-2023-geneva}.

We initially drafted event-specific roles through a crowdsourced survey with 100 participants.
Through manual inspection, we extract the frequently mentioned roles in the responses. 
These are augmented with more typical roles like \textit{Time} and \textit{Place}.
We further expand the ontology with epidemic-specific roles (e.g. \textit{Effectiveness of a cure}, \textit{Duration of a symptom}) from a fine-grained COVID ontology ExcavatorCovid \cite{min2021excavatorcovid}.
Finally, the roles are renamed to reflect corresponding events (e.g. the person who gets infected in the \textit{infect} event is named \textit{Infected}).
This multi-perspective role curation approach enhances the diversity and coverage of our ontology.

\paragraph{Filtering and Validation}
To ensure the relevance of our ontology for social media, we analyzed the event roles based on their frequency on Twitter.
Specifically, we sampled 50 tweets from the SPEED dataset and annotated them with our event roles.
Based on this analysis, we filtered out event roles with too few occurrences, such as \textit{Origin} (the source of the disease) and \textit{Manner} (how a person was infected).
Additionally, we merged roles that were too similar (e.g. \textit{Impact} and \textit{Effectiveness} are merged).
Finally, we validate our ontology with two public health experts (epidemiologists from the Dept. of Public Health).
The final ontology of event types and roles is presented in Table~\ref{tab:ontology} with definitions and examples in Appendix \S~\ref{sec:appendix-ontology}.

\begin{table}[t]
    \centering
    \small
    \begin{tabular}{l|p{5cm}}
        \toprule
        \textbf{Event Type} & \textbf{Argument Roles} \\
        \midrule
        Infect & infected, disease, place, time, value, information-source\\
        \midrule
        Spread & population, disease, place, time, value, information-source, trend\\
        \midrule
        Symptom & person, symptom, disease, place, time, duration, information-source\\
        \midrule
        Prevent & agent, disease, means, information-source, target, effectiveness\\
        \midrule
        Control & authority, disease, means, place, time, information-source, subject, effectiveness\\
        \midrule
        Cure & cured, disease, means, place, time, value, facility, information-source, effectiveness, duration\\
        \midrule
        Death & dead, disease, place, time, value, information-source, trend\\
        \bottomrule
    \end{tabular}
    \caption{Event Ontology for \dataName{} comprising 7 event types and 20 argument roles.}
    \label{tab:ontology}
\end{table}

\begin{table*}[t]
    \centering
    \small
    \begin{tabular}{lccccccccc}
        \toprule
        \multirow{2}{*}{\textbf{Dataset}} & \multirow{2}{*}{\textbf{\# Langs}} & \textbf{\# Event} & \textbf{\# Arg} & \multirow{2}{*}{\textbf{\# Sent}} & \multirow{2}{*}{\textbf{\# EM}} & \textbf{Avg. EM} & \multirow{2}{*}{\textbf{\# Args}} & \textbf{Avg. Args} & \multirow{2}{*}{\textbf{Domain}} \\
         & & \textbf{Types} & \textbf{Roles} & & & \textbf{per Event} & & \textbf{per Role} \\
        \midrule
        \textbf{Genia2013} & 1 & 13 & 7 & 664 & 6,001 & 429 & 5,660 & 809 & Biomedical \\
        \textbf{MLEE} & 1 & 29 & 14 & 286 & 6,575 & 227 & 5,958 & 426 & Biomedical \\
        % \midrule
        \textbf{ACE} & 3 & 33 & 22 & 29,483 & 5,055 & 153 & 15,328 & 697 & News \\
        \textbf{ERE} & 2 & 38 & 21 & 17,108 & 7,284 & 192 & 15,584 & 742 & News \\
        % \textbf{M}$^2$\textbf{E}$^2$ & 8 & 15 & 6,013 & $1,105$ & $138$ & News \\
        \textbf{MEE} & 8 & 16 & 23 & 31,226 & 50,011 & 3126 & 38,748 & 1685 & Wikipedia \\
        \textbf{\dataName} & 4 & 7 & 20 & 5,107 & 4,677 & 668 & 13,827 & 691 & Social Media \\
        \bottomrule
    \end{tabular}
    \caption{Data statistics for \dataName{} dataset and comparison with other standard EE datasets. Langs = languages, \# = number of, Avg. = average, Sent = sentences, EM = event mentions, Args = arguments.}
    \label{tab:main-data-statistics}
\end{table*}

\subsection{Data Processing}
\label{sec:data-processing}

We utilize Twitter as the social media platform and focus on four diseases - COVID-19, Monkeypox (MPox), Zika, and Dengue.
To maintain a similar distribution, we follow the data processing process from SPEED \cite{speed}.
For English, we directly utilize the base data provided by SPEED which dated tweets from May 15 to May 31, 2020.
For other languages, we extract tweets in the same date range as SPEED utilizing Twitter COVID-19 Endpoint as the COVID-19 base dataset.
We utilize dumps from \citet{Dias2020} as the Zika+Dengue base dataset.
% Our primary dataset consisted of a randomized selection of 331 million tweets collected between May 15 and May 31, 2020.
For tweet preprocessing, we follow \citet{DBLP:journals/eswa/PotaVFE21}: (1) anonymizing personal information, (2) normalizing retweets and URLs, and (3) removing emojis and segmenting hashtags.

\paragraph{Event-based Filtering}
To reduce annotation costs, we utilize SPEED's event filtering technique.
% hich employs sentence-level similarity measures.
Specifically, each event type is associated with a seed repository of 5-10 tweets in each language.
Query tweets are filtered based on their similarity to this seed repository.
For procuring the multilingual event-specific seed sentences, we translate the original English seed tweets into different languages.
% \jeffrey{should we talk about why we translate instead of coming up with new seed tweets or is this fine without justification}
To improve filtering efficiency, we additionally conduct keyword-based filtering for specific language-event pairs (e.g. Japanese-symptom, Japanese-cure, etc.).
Here, we filtered out a query tweet if it did not contain any event-specific keywords.
% \tanmay{@sreya Add the event name in the example}
% mere sentence-level similarity was not efficient enough for filtering.
% We supplemented this with additional keyword-based filtering wherein we filtered out a query tweet if it didn't comprise any event-specific keywords.
Finally, we apply event-based sampling from SPEED to procure the final base dataset that is utilized for data annotation.
Additional details are discussed in \S~\ref{sec:appendix-data-processing}.
% Most tweets in our primary dataset expressed subjective sentiments, with only 3\% referencing events aligned with our ontology. To reduce annotation costs, we further filtered these tweets using a simple sentence embedding similarity technique. Each event type was associated with a seed repository of 5-10 diverse tweets. Query tweets were filtered based on their sentence-level similarity to this event-based seed repository, effectively filtering out about 95% of tweets from our primary dataset, resulting in a 20-fold reduction.

% Multilingual Event-argument based filtering:
% Seed tweet based filtering: Using a set of seed tweets for languages(Japanese, Spanish, and Hindi) for each event and argument, we were able to use similarity of the seed tweets for the whole database of tweets to extract similar tweets. One way of creating the seed tweets was through skimming through the datasets and having 5-7 per event. Another way was through translating English seed tweets into the respective language, which turned out to be better for sampling based on similarity.  
% Keyword-based Filtering: After using seed tweet based filtering, we found that for some languages and events simply using seed tweets wasn’t enough to extract the tweets. For example, in Japanese \& (insert event), we needed to search for tweets that had a specific keyword in order to have a better accuracy with getting the tweets. The keyword-based filtering essentially enhanced the seed tweet based filtering methodology. 

\subsection{Data Annotation}

We conduct two sets of annotations to create our multilingual EE dataset: (1) EAE annotations for existing SPEED English ED data and (2) ED+EAE annotations for data in Japanese, Hindi, and Spanish.
For ED, annotators were tasked to identify the presence of any events in a given tweet.
For EAE, annotators were further asked to identify and extract event-specific roles that were also mentioned in the tweet.
We provide further details about the annotation guidelines in \S~\ref{sec:appendix-data-annotation}.

\paragraph{Annotators and Agreement}
To maintain high annotation quality, our annotators were selected to be a pool of seven experts who were computer science NLP students trained through multiple annotation rounds.
Of these seven, we had three annotators who were bilingual speakers of English and Japanese/Hindi/Spanish respectively.
These three annotators handled the multilingual ED and EAE annotations.
The remaining four annotators, along with the bilingual English-Hindi annotator, focused on English EAE annotations.
% Three experts are bilingual in Japanese/Hindi/Spanish and English and handled the multilingual annotations while the rest focused on English EAE.

To ensure good annotation agreement, we conduct two agreement studies among the annotators: (1) ED annotations for multilingual annotators and (2) EAE annotations for all annotators.
Both these studies were conducted using English data (even for multilingual annotations) to ensure that agreement could be measured in a fair manner.
Agreement scores were measured using Fleiss' Kappa \cite{fleiss1971measuring}.
For ED agreement, two rounds of study for the 3 multilingual annotators yielded a super-strong agreement score of 0.75 (30 samples).
For EAE, the agreement score for the 7 annotators after two annotation rounds was a decent 0.6 (25 samples).

\paragraph{Annotation Verification}
To mitigate single annotator bias, each datapoint in the English data is annotated by two annotators, with a third annotator resolving inconsistencies.
Owing to the scarcity of multilingual annotators, we hire three additional bilingual speakers to verify the multilingual annotations.
These verification annotators were selected through a thorough qualification test to ensure high verification quality.
They were requested to judge if the current annotations were reasonable.
% We validate their agreement with agreement score of 0.6.
If the original annotation was deemed incorrect, they were asked to provide feedback to correct the annotations.
This feedback was finally utilized by the original multilingual annotators to rectify the annotation.
We provide additional details in \S~\ref{sec:appendix-multilingual-data-verification}.

\subsection{Data Analysis}

\paragraph{Comparison with other datasets}
\dataName{} comprises 5,106 tweets with 4,674 event mentions and 13,815 argument mentions across four diseases and four languages.
We present the main statistics along with comparisons with other prominent EE datasets like ACE \cite{doddington-etal-2004-automatic}, ERE \cite{song-etal-2015-light}, Genia2013 \cite{li-etal-2020-cross}, MEE \cite{pouran-ben-veyseh-etal-2022-mee}, and MLEE \cite{DBLP:journals/bioinformatics/PyysaloOMCTA12} in Table~\ref{tab:main-data-statistics}.
We note that \dataName{} is one of the few multilingual EE datasets, notably the first in the social media domain.
Overall, \dataName{} is comparable in various event and argument-related statistics with the previous standard EE datasets.

\begin{table}[t]
    \centering
    \small
    \begin{tabular}{lcccc}
        \toprule
        \textbf{Lang} & \textbf{\# Sent} & \textbf{Avg. Length} &  \textbf{\# EM} & \textbf{\# Args} \\
        \midrule
        en & 2,560 & 32.5 & 2,887 & 8,423 \\
        es & 1,012 & 32.4 & 614 & 1,485 \\
        hi & 716 & 30.0 & 627 & 2,344 \\
        ja & 819 & 89.2* & 549 & 1,575 \\
        \bottomrule
    \end{tabular}
    \caption{Data statistics for \dataName{} split by language. \# = number of, Avg = average, Lang = language, Sent = sentences, Args = arguments, *character-level.}
    \label{tab:multi-data-statistics}
\end{table}

\paragraph{Multilingual Statistics}
We provide a deeper split of data statistics per language in Table~\ref{tab:multi-data-statistics}.
Owing to cheaper annotations, English has many more annotated sentences compared to other languages.
This is also a design choice, as we will solely utilize English data for training zero-shot multilingual models (discussed in \S~\ref{sec:epidemic-ee}).
In terms of event and argument densities (i.e. \# EM / \# Sent and \# Args / \# Sent), we notice a broader variation across languages, with English and Hindi being denser.
The average lengths (in terms of the number of words) are similar across the languages.

\begin{figure}[t]
    \centering
    \includegraphics[width=0.48\textwidth]{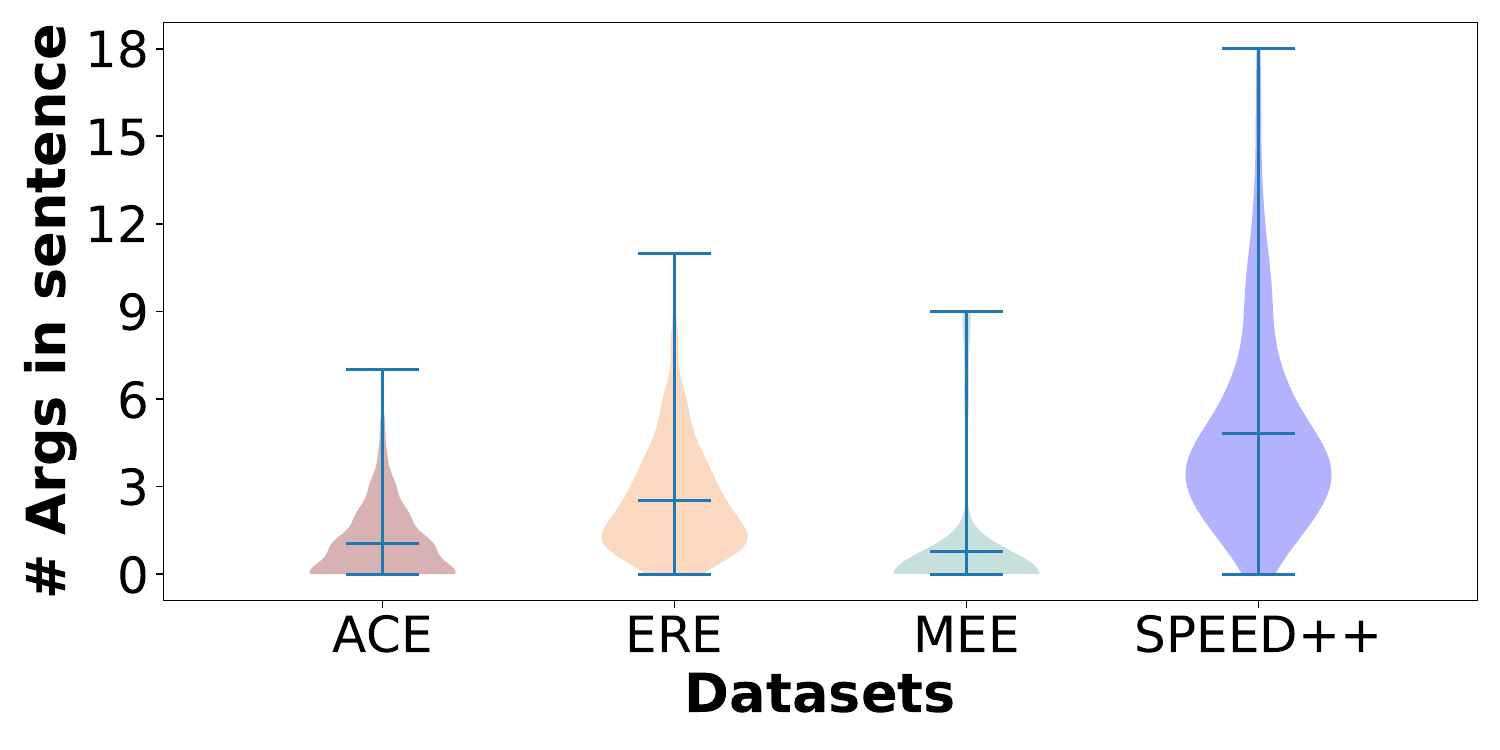}
    \caption{Distribution of the number of arguments (\# Args) per sentence for \dataName{} relative to other datasets ACE, ERE, and MEE.}
    \label{fig:arg-density}
\end{figure}

\paragraph{Argument Study}
We deep-dive to study the density in terms of arguments per sentence for \dataName{} by comparing with standard EE datasets ACE and ERE and a multilingual EE dataset MEE in Figure~\ref{fig:arg-density}.
Noticeably, \dataName{} is more dense (has a higher mean argument per sentence value) and has a broader distribution with sentences up to 18 arguments as well.
Furthermore, following GENEVA \cite{parekh-etal-2023-geneva}, we add 4 non-entity roles, which make up 20\% of the total arguments.
Such non-entity arguments are not present in any other multilingual datasets.
Overall, the high and broad argument density and the existence of non-entity arguments render \dataName{} to be a more challenging EE dataset.

\section{Zero-shot Cross-lingual Event Extraction}
\label{sec:epidemic-ee}

\begin{table}[t]
    \centering
    \small
    \begin{tabular}{l|rrrr}
        \toprule
        & \textbf{Disease} & \textbf{Language} & \textbf{\# Sent} & \textbf{\# EM}\\
        \midrule
        \textbf{Train} & COVID & English & 1,601 & 1,746 \\ \midrule
        \textbf{Dev} & COVID & English & 374 & 471 \\ \midrule
        \multirow{8}{*}{\textbf{Test}} & \multirow{3}{*}{COVID} & Spanish & 534 & 365 \\
        & & Hindi & 416 & 412 \\
        & & Japanese & 542 & 395 \\ \cline{2-5}
        & Monkeypox & English & 286 & 398 \\ \cline{2-5}
        & \multirow{4}{*}{Zika + Dengue} & English & 299 & 272 \\
        & & Spanish & 478 & 249 \\
        & & Hindi & 300 & 215 \\
        & & Japanese & 277 & 154 \\
        \bottomrule
    \end{tabular}
    \caption{Data split for epidemic event extraction. \# = number of, Sent = sentences, EM = event mentions.}
    \label{tab:data-setup}
\end{table}

\begin{table*}[t]
    \centering
    \small
    \setlength{\tabcolsep}{2.6pt}
    \begin{tabular}{l|cc|cc|cc|cc|cc|cc|cc|cc|cc}
        \toprule
        \multirow{3}{*}{\textbf{Model}} & \multicolumn{6}{c|}{\textbf{COVID}} & \multicolumn{2}{c|}{\textbf{MPox}} & \multicolumn{8}{c|}{\textbf{Zika + Dengue}} & \multicolumn{2}{c}{\textbf{Average}} \\
        & \multicolumn{2}{c}{hi} & \multicolumn{2}{c}{jp} & \multicolumn{2}{c|}{es} & \multicolumn{2}{c|}{en} & \multicolumn{2}{c}{en} & \multicolumn{2}{c}{hi} & \multicolumn{2}{c}{jp} & \multicolumn{2}{c|}{es} \\ \cmidrule(r){2-19}
        & \textbf{EC} & \textbf{AC} & \textbf{EC} & \textbf{AC} & \textbf{EC} & \textbf{AC} & \textbf{EC} & \textbf{AC} & \textbf{EC} & \textbf{AC} & \textbf{EC} & \textbf{AC} & \textbf{EC} & \textbf{AC} & \textbf{EC} & \textbf{AC} & \textbf{EC} & \textbf{AC} \\
        \midrule
        \multicolumn{19}{c}{\small \textsc{Baseline Models}} \\
        \midrule
        ACE - TagPrime & 0.0 & 0.0 & 0.0 & 0.0 & 0.0 & 0.0 & 0.0 & 0.0 & 0.0 & 0.0 & 0.0 & 0.0 & 0.0 & 0.0 & 0.0 & 0.0 & 0.0 & 0.0\\
        % MEE - TagPrime & & & & & & & & & & & & & & & & \\
        DivED$^*$$^\ddagger$ & 0.0 & - & 0.0 & - & 27.0 & - & 36.7 & - & 47.7 & - & 4.4 & - & 1.3 & - & 12.8 & - & 16.2 & - \\
        Keyword$^\star$ & 15.2 & - & 28.6 & - & 26.3 & - & 41.3 & - & 39.3 & - & 18.6 & - & \textbf{39.7} & - & 20.6 & - & 28.7 & - \\
        COVIDKB$^\dagger$ & 45.2 & - & 42.3 & - & 24.2 & - & 18.5 & - & 45.5 & - & 47.5 & - & 34.0 & - & 34.6 & - & 36.5 & - \\
        GPT-3.5-turbo$^\star$ & 35.7 & 14.5 & 36.4 & 15.0 & 43.2 & 16.8 & 46.4 & 24.0 & 56.6 & 33.0 & 45.5 & 20.0 & 29.0 & 11.0 & 39.6 & 15.1 & 41.6 & 18.7 \\
        \midrule
        \multicolumn{19}{c}{\small \textsc{Trained on SPEED++ (Our Framework)}} \\
        \midrule
        TagPrime & 60.1 & \textbf{39.0} & 35.3 & 7.9 & 62.0 & 37.1 & \textbf{70.2} & \textbf{45.1} & \textbf{66.7} & \textbf{48.5} & 65.0 & \textbf{40.7} & 27.2 & 7.5 & \textbf{49.9} & \textbf{28.2} & \underline{54.6} & \underline{31.8} \\
        TagPrime + XGear & 60.1 & 27.1 & 35.3 & 9.7 & 62.0 & 36.0 & \textbf{70.2} & 42.0 & \textbf{66.7} & 45.8 & 65.0 & 31.9 & 27.2 & 8.8 & \textbf{49.9} & 26.6 & 54.6 & 28.5 \\
        BERT-QA & 54.7 & 33.9 & 21.2 & 4.0 & 60.6 & 28.1 & 66.1 & 39.0 & 63.0 & 45.5 & 50.8 & 31.1 & 4.6 & 0.8 & 41.9 & 24.1 & 45.4 & 25.8 \\
        DyGIE++ & 61.0 & 35.7 & 38.2 & 2.0 & 61.7 & \textbf{39.1} & 67.4 & 39.3 & 64.1 & 46.4 & 61.4 & 32.2 & 27.5 & 0.4 & 45.6 & 22.7 & 53.4 & 27.2 \\
        OneIE & \textbf{61.9} & 34.3 & 12.0 & 11.4 & 44.5 & 37.3 & 68.8 & 42.6 & \textbf{66.7} & 47.9 & 61.9 & 38.5 & 12.0 & 5.0 & 44.5 & 25.4 & 46.5 & 30.3 \\
        \midrule
        % \multicolumn{17}{c}{\small \textsc{Trained on SPEED++ (Our Framework) with CLaP}} \\
        % \midrule
        TagPrime + CLaP & 58.6 & 32.9 & \textbf{48.4} & \textbf{19.1} & \textbf{62.6} & 37.7 & \textbf{70.2} & \textbf{45.1} & \textbf{66.7} & \textbf{48.5} & \textbf{65.2} & 40.6 & 39.2 & \textbf{18.8} & 49.7 & 28.1 & \textbf{57.6} & \textbf{33.9} \\
        % TagPrime ED + XGear EAE & & & & & & & & & & & & & & & & \\
        % EEQA ED + EAE & & & & & & & & & & & & & & & & \\
        
        \bottomrule
    \end{tabular}
    \caption{Benchmarking EE models trained on \dataName{} for extracting event information in the cross-lingual cross-disease setting. EC = Event Classification, AC = Argument Classification, hi = Hindi, jp = Japanese, es = Spanish, and en = English. $^\star$Numbers are higher due to string matching evaluation. $^\dagger$Binary classification evaluation.}
    \label{tab:ec-main-results}
\end{table*}

To validate the effectiveness of EE for epidemic events, we benchmark various EE models using \dataName.
Given the infeasibility of procuring quality data in all languages for all diseases, we benchmark in a zero-shot cross-lingual cross-disease fashion i.e. we train models only on English COVID data and evaluate on the rest.
We provide the data split for our benchmarking in Table~\ref{tab:data-setup}.
For evaluation, we report F1-score for event classification (\textbf{EC}) and argument classification (\textbf{AC}) \cite{ahn-2006-stages} measuring the classification of events and arguments respectively.
We use TextEE \cite{huang2024textee} for most implementations, with specific details discussed in \S~\ref{sec:appendix-implementation-details}. 
Additional benchmarking experiments are provided in \S~\ref{sec:additional-eee-expts}.
\paragraph{EE Models}
% \begin{enumerate}
    % \item Discuss two ways to do EE - end2end and pipelined. Discuss the various models we are benchmarking and their alliance with both these paradigms.
% \end{enumerate}
% Since our focus is on cross-lingual EE, many of the English-based EE models can't be used for our setting.

% Most EE models solely focus on English and can't be directly utilized in the cross-lingual setting.
% To this end, we adapt the following models using multilingual pre-trained models and tokenization:
% (1) TagPrime \cite{hsu-etal-2023-simple}, a sequence tagger priming words to input text to convey more task-specific information.
% (2) EEQA \cite{du-cardie-2020-event}, a classification model utilizing label semantics by formulating event extraction as a question-answering task.
% (3) DyGIE++ \cite{wadden-etal-2019-entity}, a multi-task end-to-end classification-based model using span graph propagation to aggregate local and global context. 
% (4) OneIE \cite{lin-etal-2020-joint}, a joint-training model training a single model for various information extraction tasks.
% (5) XGear \cite{huang-etal-2022-multilingual-generative}, a language-agnostic generative EAE model which applies generative methods for EE. Since XGear is an EAE-only model, we combine it with the ED outputs from TagPrime.
% To further improve the models, we train them with augmented pseudo-generated data using a label projection model CLaP \cite{clap}.

Most EE models solely focus on English and can not be directly utilized in the cross-lingual setting.
To this end, we adopt the following models using multilingual pre-trained models and tokenization:
(1) TagPrime \cite{hsu-etal-2023-tagprime},
% a sequence tagger priming words to input text to convey more task-specific information.
(2) EEQA \cite{du-cardie-2020-event},
% a classification model utilizing label semantics by formulating event extraction as a question-answering task.
(3) DyGIE++ \cite{wadden-etal-2019-entity},
% a multi-task end-to-end classification-based model using span graph propagation to aggregate local and global context. 
(4) OneIE \cite{lin-etal-2020-joint},
% a joint-training model training a single model for various information extraction tasks.
(5) XGear \cite{huang-etal-2022-multilingual-generative}.
% a language-agnostic generative EAE model which applies generative methods for EE.
Since XGear is an EAE-only model, we combine it with the TagPrime ED model.
To further improve the models, we train them with pseudo-generated data using a label projection model CLaP \cite{clap}.

\paragraph{Baseline Models}
% \begin{enumerate}
    % \item Check if any zero-shot EE models can be tested here - maybe Po-Nien's / Derek's work?
    % \item LLM/GPT baseline
    % \item Data Transfer models / COVIDKB model?
% \end{enumerate}
We consider the following baselines:
(1) ACE - TagPrime, a TagPrime model trained on the multilingual EE dataset ACE \cite{doddington-etal-2004-automatic}
% and MEE \cite{pouran-ben-veyseh-etal-2022-mee}
and transferred to \dataName{}.
(2) DivED \cite{DBLP:journals/corr/abs-2403-02586}, a Llama2-7B model fine-tuned on a diverse range of event definitions,
(3) COVIDKB \cite{zong-etal-2022-extracting}, an epidemiological work using a BERT classification model. Since the original output classes are different, we train it to simply classify tweets as epidemic-related or not.
(4) Keyword baseline inspired from an epidemiological work \cite{DBLP:journals/artmed/LejeuneBDL15} curates a set of keywords for each event.
% We curate keywords for English and translate them into other languages for this baseline.
(5) GPT-3.5-turbo \cite{DBLP:journals/corr/abs-2005-14165}, a Large Language Model (LLM) baseline using seven in-context examples.
% \tanmay{Check if any zero-shot EE models can be tested here - maybe Po-Nien's / Derek's work?}

\paragraph{Results}
We present our per-disease per-language results in Table~\ref{tab:ec-main-results}.
We note that most of the baseline models do not perform well for our task, as was noted also in \citet{speed}.
The GPT-based LLM baseline performs better in English but exhibits poor performance across other languages.
% \jeffrey{do we use present or past tense in this section}
On the other hand, we observe stronger performance by the \textbf{supervised baselines trained on our \dataName{} dataset with TagPrime providing the best overall average performance}.
We also note how CLaP further improves performance by 2-3 F1 points in the cross-lingual setting, especially for character-based language Japanese.
\blfootnote{$^\ddagger$English-based Llama performs poorly multilingually.}

% \begin{enumerate}
%     \item Discuss comparison of supervised SPEED-trained models with others
%     \item Quick discussion of end2end and pipelined models
% \end{enumerate}

% \subsection{Multilingual EE}

% \tanmay{@hyosang}
% \paragraph{Multilingual EE Models}
% \begin{enumerate}
%     \item Discuss the models that allow multilingual EE - TagPrime, XGear... can we also try EEQA here?
%     \item Discuss CLaP method
% \end{enumerate}

% \paragraph{Baseline Models}
% \begin{enumerate}
%     \item Check if any zero-shot EE models can be tested here - maybe Po-Nien's / Derek's work?
%     \item LLM/GPT baseline
% \end{enumerate}

% \paragraph{Results}
% \begin{enumerate}
%     \item Discuss poor transfer to other languages - important benchmark to continue to improve zero-shot cross-lingual transfer
% \end{enumerate}

% \subsection{Conclusions}
% Discuss how we use these models for future applications

\section{Applications}

To validate its practical utility for epidemic preparedness, we demonstrate our framework's use in two downstream applications: Global Epidemic Prediction and Epidemic Information Aggregation.
For this, we train a multilingual TagPrime model on the entire \dataName{} dataset.
Further details about these applications are provided below.

\subsection{Global Epidemic Prediction}
\label{sec:global-epidemic-prediction}

% Local infections and diseases continuously spread worldwide at all times and any such disease spread can constitute an epidemic \cite{morse2001factors}.
% Naturally, people start discussing local epidemics in regional languages - which are not English 80\% of the time.
% Thus, there is an underlying need to detect such epidemic discussions multilingually for earlier global epidemic prediction.
To showcase the robust multilingual utility of our framework, we highlight its extensive language coverage and provide an in-depth analysis of COVID-19 predictions from Chinese data.

\begin{figure}[t]
    \centering
    \includegraphics[width=0.48\textwidth]{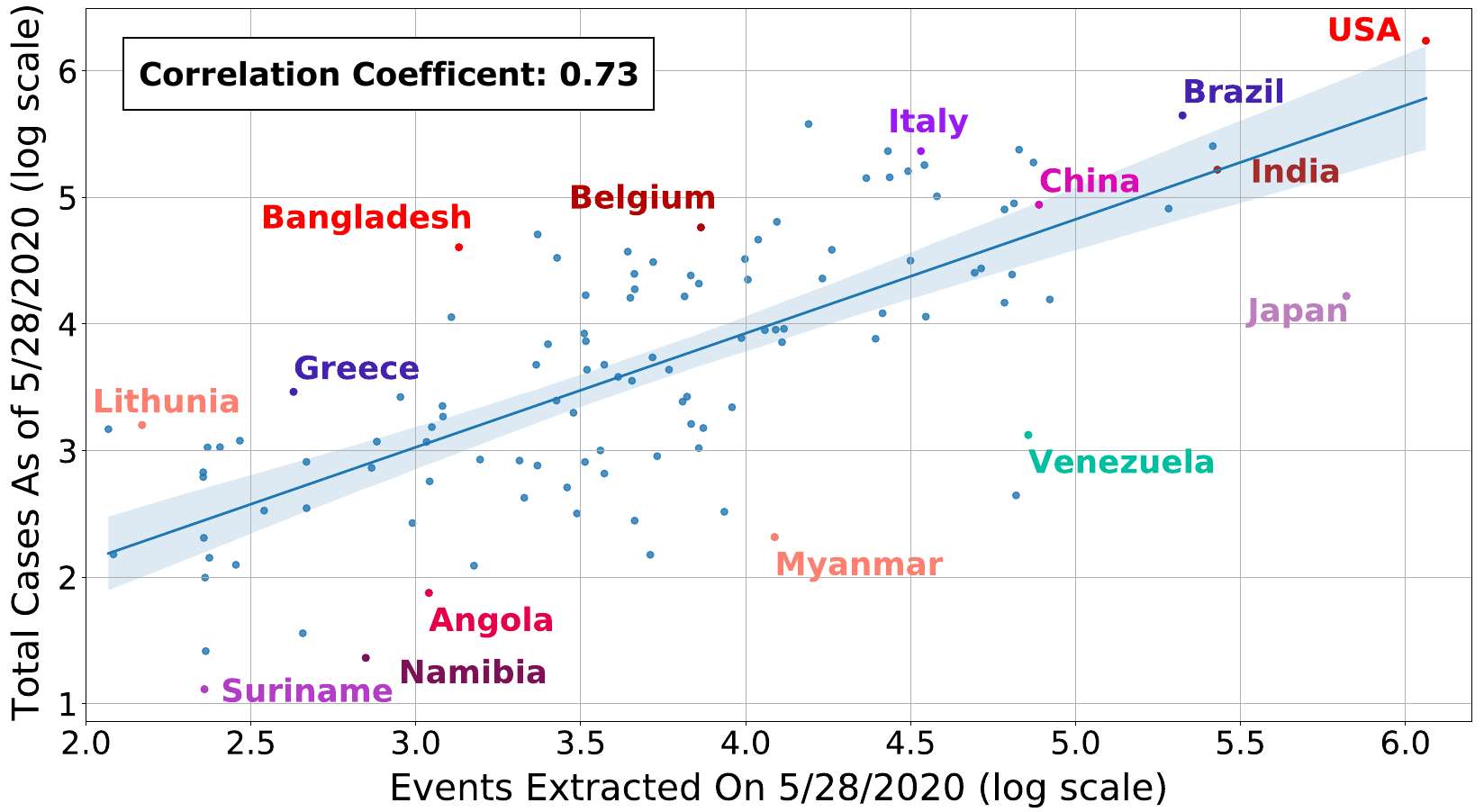}
    \caption{Number of extracted events plotted against the number of reported cases for each country. Both of them are in log scale.}
    \label{fig:events-vs-cases}
\end{figure}

\begin{figure}[t]
    \centering
    \includegraphics[width=0.45\textwidth]{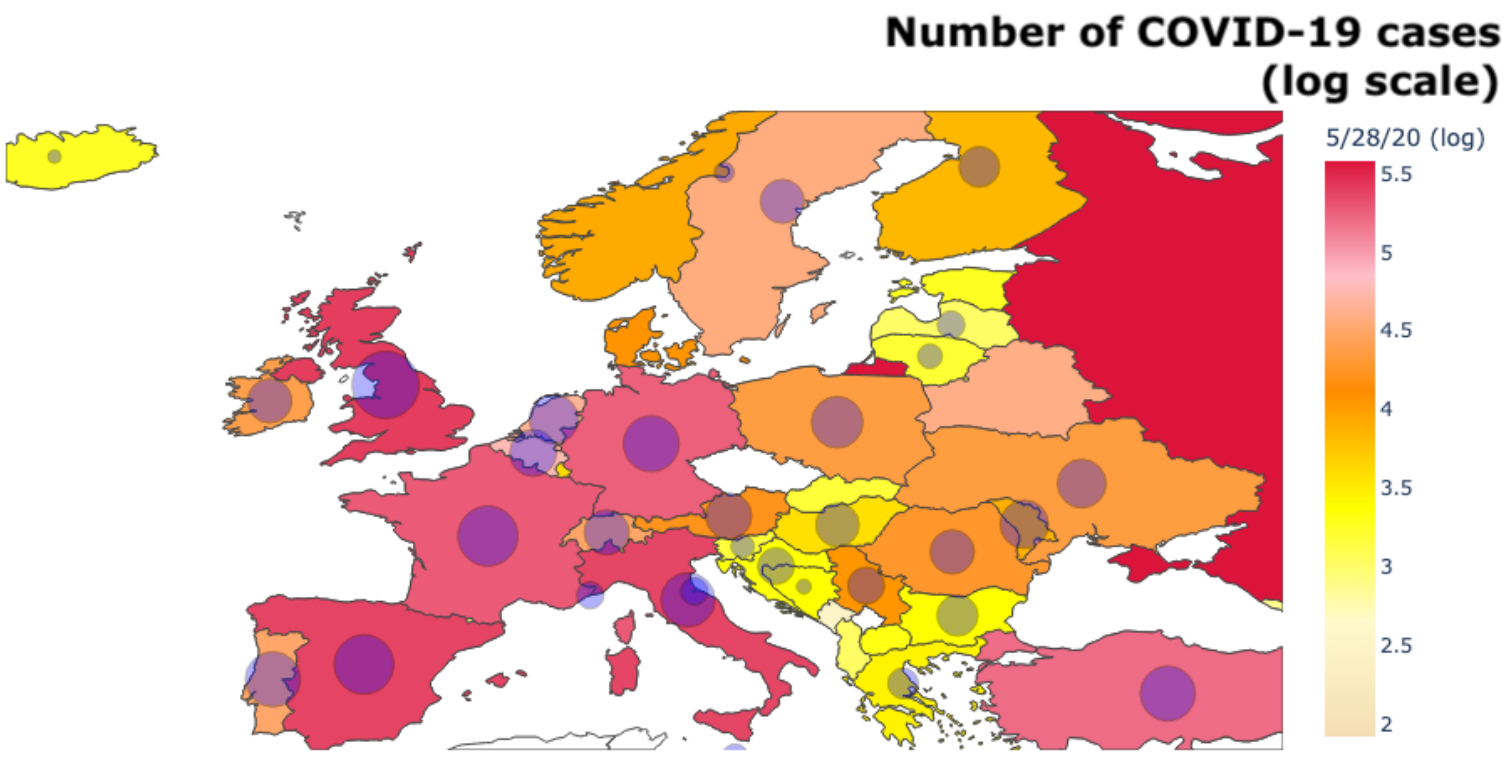}
    \caption{Geographical distribution of the number of reported COVID-19 cases as of May 28, 2020 in Europe. Red depicts more spread, and yellow/white indicates less spread. The blue dots indicate the events extracted by our model and its size depicts the number of epidemic events for the specific country (log scale).}
    \label{fig:europe-map-plot}
\end{figure}

\paragraph{Global Epidemic Monitoring}
Validating our framework for each language is resource-intensive and infeasible.
Instead, we perform a preliminary study of the broad language coverage of our framework by demonstrating its capability to detect COVID-related events across 65 languages.
We analyze tweets across all languages from a single day (May 28, 2020) and extract epidemic events using our framework.
Utilizing user locations, we map the tweets from these languages to 117 countries.
For reference, we plot the extracted events for each country with the actual number of reported COVID-19 cases\footnote{\url{https://www.worldometers.info/coronavirus/}} in Figure~\ref{fig:events-vs-cases}.
Countries with significant COVID-19 spread appear in the top-right, while some outliers are also shown in the figure.
Our framework achieves a healthy correlation of 0.73 with the actual reported cases, indicating strong performance across a broad range of languages.
% \footnote{In comparison, the total number of posts achieves a correlation of 0.65 with the reported number of cases.}

We further extend these plots geographically in Figure~\ref{fig:europe-map-plot}.
Each country is color-coded on the number of COVID cases, with lighter shades indicating fewer cases while darker shades indicate a massive spread.
The extracted number of events from the country-mapped tweets by our framework are plotted as translucent circles.
Bigger dots indicate more events extracted for the given country.
In this plot, we observe extracted events and COVID-19 spread more in Western European countries like the United Kingdom, France, Spain, Italy, and Germany, while lesser events spread in Eastern European countries.
We provide additional details along with a world map geographical plot in \S~\ref{sec:appendix-gep-additional-details}.

\paragraph{COVID-19 Epidemic Prediction using Chinese}
As a case study, we examine the earliest stages of the COVID-19 pandemic, analyzing Chinese social media posts from Dec 16, 2019, to Jan 21, 2020, using Weibo data from \citet{hu-etal-2020-weibo}.
Using our trained TagPrime model, we infer on Chinese in a \textit{zero-shot fashion} (i.e. without prior training on Chinese).
We aggregate the 7-day rolling average of our extracted event mentions across time and report any sharp increases as epidemic warnings, as illustrated in Figure~\ref{fig:zh-timeseries}.
Since case reporting had not begun, actual COVID-19 case numbers are unavailable for this period.
Instead, we also plot the total number of Weibo posts and active users \cite{Guo2021ImprovingGF}.
Additionally, we compare trends with baselines from COVIDKB \cite{zong-etal-2022-extracting} and a keyword-based approach \cite{DBLP:journals/artmed/LejeuneBDL15}.
% \tanmay{Clarify why not reporting cases}
% \tanmay{linebreak and arrow in future for CDC}

\begin{table}[t]
    \centering
    \small
    \begin{tabular}{p{2.6cm}|p{4.25cm}}
        \toprule
        \textbf{Chinese Posts} & \textbf{Translation} \\
        \midrule
        \zh{武汉华南海鲜市场\bluetext{出现 [infect]}多个不明原因肺炎病例，请同道们提高警惕，早期发现，早期\bluetext{隔离 [prevent]}} & Multiple cases of pneumonia of unknown origin have \bluetext{appeared [infect]} in Wuhan's Huanan Seafood Market. Please be more vigilant, detect and \bluetext{isolate [prevent]} them as early as possible. \\
        \midrule
        \zh{近日，武汉进入流感\bluetext{高发期 [spread]}，多家医院\bluetext{感冒 [symptom]}发烧的患者数量猛增。} & Recently, Wuhan has entered a period of high influenza \bluetext{incidence [spread]}, and the number of patients with \bluetext{colds [symptom]} and fevers in many hospitals has increased sharply. \\
        \bottomrule
    \end{tabular}
    \caption{Sample Weibo posts in Chinese with their translations identified by \dataName{} framework as epidemic-related from late December 2019. Event types and their trigger words are marked in \bluetext{blue}.}
    \label{tab:chinese-example-posts}
\end{table}

\begin{figure*}
    \centering
    \includegraphics[width=0.98\textwidth]{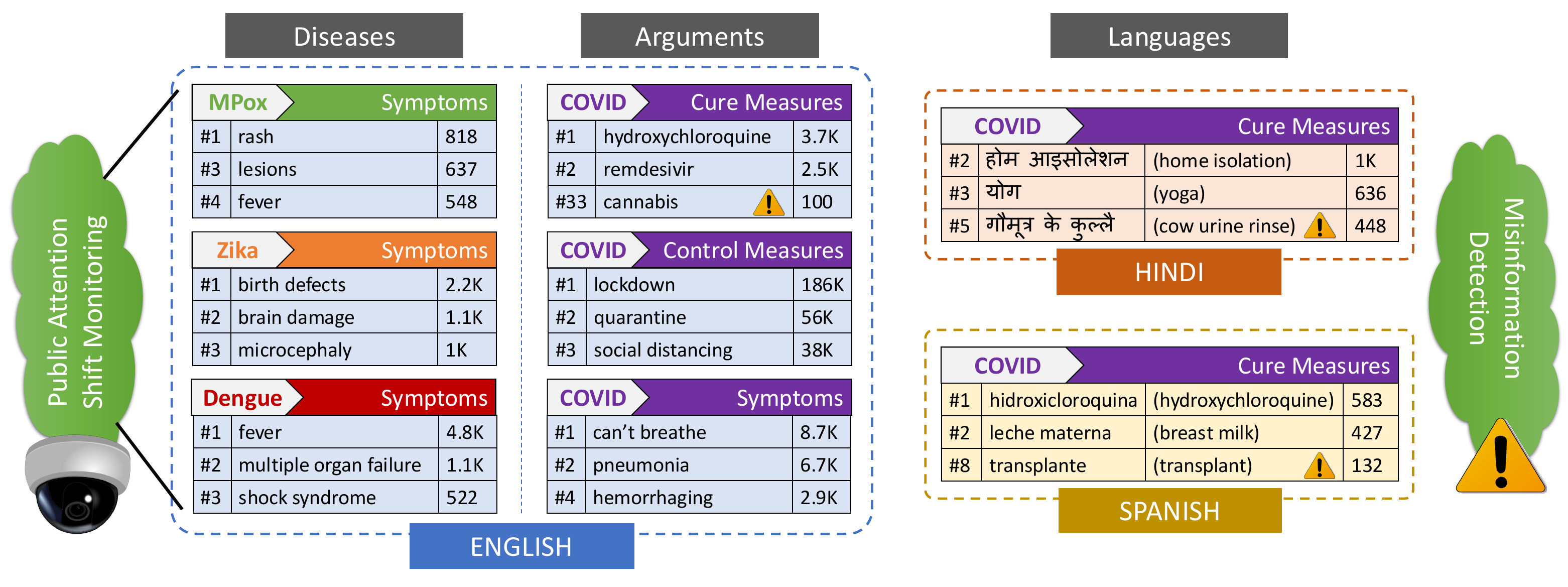}
    \caption{Information Assimilation Bulletin as extracted by our \dataName{} framework and agglutinatively clustered. The first column represents different diseases, the second column represents different argument roles, and the third column represents the different languages. We also highlight the utility of these bulletin for two applications of Public Attention Shift Monitoring and Misinformation Detection.}
    \label{fig:bulletin}
\end{figure*}

Figure~\ref{fig:zh-timeseries} demonstrates how our \dataName{} framework provides epidemic warnings three weeks before the global tracking of infection cases began.
While the keyword-based method also provides some signals, they are relatively weaker.
Furthermore, Table~\ref{tab:ec-main-results} shows that the keyword baseline performs worse for morphologically richer languages, making it less robust.
Additionally, the number of posts and active users do not provide any epidemic-related signals.
For further validation, we present sample event mentions extracted by our framework in Table~\ref{tab:chinese-example-posts}.
In the bottom timeline\footnote{\url{https://www.cdc.gov/museum/timeline/covid19.html}} in Figure~\ref{fig:zh-timeseries}, we demonstrate the efficacy of our framework as it provided epidemic warnings 6 weeks before the "COVID-19" term was coined and used in social media.
Overall, we show how \textbf{our framework can provide early epidemic warnings multilingually without relying on any target language data, making it suitable for global deployment.}

% Diagram - Europe Map Severity

% \includegraphics[width=0.47\textwidth]{figures/Europe_Graph.png}

% Diagram - World Map Severity

% \includegraphics[width=0.47\textwidth]{figures/Cases_Events.png}

\subsection{Epidemic Information Aggregation}
\label{sec:epidemic-info-assimilation}

Our framework possesses a strong EAE capability to extract detailed information about epidemic events such as \textit{symptoms}, \textit{preventive measures}, \textit{cure measures}, etc.
Aggregating such information from millions of social media posts can provide insights into public opinions regarding various epidemic aspects.
% Extracting and aggregating such information from millions of social media posts can provide insights about general public opinions regarding various aspects of the epidemic.
To this end, we develop an information aggregation system for community epidemic discussions.
Specifically, we use our framework to extract arguments for various event roles, project them into a representative space, and merge similar arguments using agglomerative clustering.
The final arguments and their counts for different diseases and languages, extracted from 6M tweets, are presented as a bulletin in Figure~\ref{fig:bulletin}.
Further details and a complete table are reported in \S~\ref{sec:appendix-eia-additional-details}.
% [Details about clustering method]
% First we project the arguments into representation space using Hugging Face SentenceTransformer distiluse-base-multilingual-cased-v2 model. Then we use sklearn's implementation of agglomerative clustering with a euclidean distance threshold of $1$ to fit the clusters. After generating the cluster, we pick the $3$ largest clusters and the argument with the most number of mentions within each cluster as the label.
% \jeffrey{@tanmay is this good}

\begin{table}[t]
    \centering
    \small
    \begin{tabular}{p{3.6cm}p{3.2cm}}
        \toprule
        \textbf{Tweet} & \textbf{Translation} \\
        \midrule
        \multicolumn{2}{c}{ \textsc{\redtext{Hindi} \quad - \quad \bluetext{COVID-19} \quad - \quad \browntext{Cure Measures}} }\\
        \midrule
        \hi{koronA k\? EKlAP ek yo\388wA bn k\? \redtext{yog} s\? koronA ko hrAnA h\4.} & Become a warrior against Corona and defeat it through \redtext{yoga} \\ \hline
        \hi{s\2Ebt koronA pA\<E)EVv{\rs ,\re} hAlt nA\7{j}k aA(mEnB\0r aEByAn k\? tht \redtext{gO\8{m}/ k\? \7{k}\3A5w\4} krvAkr upcAr EkyA jA rhA h\4.} & Sambit is corona positive, condition is critical, and is being treated by \redtext{gargling with cow urine} under the self-reliant campaign. \\ \hline
        \hi{s\2Ebt jF k\? koronA \3FEwBAEvt hon\? kF Kbr EmlF. jAnkr \7{d},K \7{h}aA{\rs ,\re} l\?Ekn v\? a-ptAl m\?{\qva} \redtext{-vA-Ly lAB} l\? rh\? h\4{\qva}{\rs ,\re} aOr ab b\7{h}t b\?htr h\4} & Received the news of Sambit ji being affected by Corona! Sorry to hear, but he's \redtext{recovering} in the hospital, and is much better now. \\
        \midrule
        \multicolumn{2}{c}{ \textsc{\redtext{Spanish} \quad - \quad \bluetext{COVID-19} \quad - \quad \browntext{Cure Measures}} }\\
        \midrule
        Científicos rusos sugieren que una proteína presente en la \redtext{leche materna} puede ser clave en la lucha contra el covid-19 (url) & Russian scientists suggest that a protein present in \redtext{breast milk} may be key in the fight against covid-19 (url) \\ \hline
        Este \redtext{transplante} de pulmones a un paciente con COVID-19 es una operación realizada hasta ahora sólo en China y que por primera vez se lleva a cabo en Europa & This lung \redtext{transplant} to a patient with COVID-19 is an operation carried out so far only in China and is being carried out for the first time in Europe. \\ \hline
        el mundo acumula más evidencia de la efectividad de la \redtext{ivermectina} para el tratamiento en casa de pacientes con estadios leves de \#(COVID)-19 & the world accumulates more evidence of the effectiveness of \redtext{ivermectin} for the home treatment of patients with mild stages of \#(COVID)-19 \\
        \bottomrule
    \end{tabular}
    \caption{Illustration of some actual tweets in Hindi and Spanish mentioning various cure measures related to COVID-19. The terms related to \textit{cure measures} extracted by our framework are highlighted in \redtext{red}.}
    \label{tab:info-assimilation-examples-multi-covid}
\end{table}

Our framework effectively extracts various arguments for COVID-19 in English (middle column of Figure~\ref{fig:bulletin}), including cure measures such as \textit{hydroxychloroquine} and \textit{remdesivir}, control measures like \textit{lockdown} and \textit{quarantine}, and symptoms such as \textit{pneumonia}.
% Across different arguments (middle column in Figure~\ref{fig:bulletin}) for COVID-19 for English, our framework is capable of extracting relevant cure measures like \textit{hydroxychloroquine} and \textit{remdesivir}, control measures like \textit{lockdown} and \textit{quarantine}, and symptoms like \textit{pneumonia}.
Additionally, this capability extends to other diseases, such as Monkeypox, Zika, and Dengue (left column), and across languages, including Hindi and Spanish (right column).
This condensed information is crucial for Public Attention Shift Monitoring, aiding policymakers in devising better control measures \cite{liu2022study}. 
% For example, different symptoms extracted for various diseases are illustrated on the left side of Figure~\ref{fig}.
% Such condensed information can be vital for \textbf{Public Attention Shift Monitoring} to aid policymakers to devise better control measures \cite{liu2022study}.
We demonstrate this in the form of extracted symptoms such as \textit{rashes} and \textit{lesions} for Monkeypox and \textit{fever} and \textit{shock syndrome} for Dengue (left column).
Simultaneously, our framework can assist in \textbf{Misinformation Detection} \cite{mendes-etal-2023-human}.
Shown as caution signs in the figure, we highlight various potential COVID-19 cure misinformation such as \textit{cow urine rinse}, \textit{cannabis}, and \textit{transplants} extracted across languages by our framework.
As evidence, we also provide some of the actual tweets in Hindi and Spanish flagged by our framework for mentioning these terms in Table~\ref{tab:info-assimilation-examples-multi-covid}.
We provide example tweets comprising these arguments as extracted by our framework in \S~\ref{sec:appendix-eia-additional-details}.
% \tanmay{populate this}

% Observations about clusters:
% \begin{enumerate}
%     \item General observations and validation that the trends are captured correctly
%     \item Misinformation detection (can have figure like previous paper in appendix)
%     \item Public sentiment monitoring (with symptom argument?)
% \end{enumerate}

% \input{sections/global_ep}

% \input{sections/info_assimilation}

\section{Related Works}
\label{sec:related-works}

\paragraph{Event Extraction Datasets}
Event Extraction (EE) aims at detecting events (Event Detection) and extracting details about specific roles associated with each event (Event Argument Extraction) from natural text.
Unlike document parsing \cite{tong-etal-2022-docee, DBLP:journals/corr/abs-2408-01046}, we utilize EE only at the sentence/tweet level in our work.
Overall, EE is a well-studied task with earliest works dating back to MUC \cite{sundheim-1992-overview, grishman-sundheim-1996-message}, ACE \cite{doddington-etal-2004-automatic} and ERE \cite{song-etal-2015-light}, but there have been various newer diverse datasets like MAVEN \cite{wang-etal-2020-maven}, WikiEvents \cite{li-etal-2021-document}, FewEvent \cite{DBLP:journals/corr/abs-1910-11621}, DocEE \cite{tong-etal-2022-docee}, and GENEVA \cite{parekh-etal-2023-geneva}.
While most of these datasets are only in English, some datasets like ACE \cite{doddington-etal-2004-automatic}, ERE \cite{song-etal-2015-light}, and MEE \cite{pouran-ben-veyseh-etal-2022-mee} provide EE data in ten languages for general-purpose events in the news and Wikipedia domains.
SPEED \cite{speed} introduces data for epidemic-based events in social media but is limited to Event Detection and focuses on English.
Overall, \dataName{} extends SPEED to four languages and Event Argument Extraction.
% \jeffrey{rewrite: \dataName{} extends SPEED to four languages and implements Event Argument Extraction on top of Event Detection}.

\paragraph{Multilingual Epidemiological Information Extraction}
Early epidemiological works \cite{lindberg1993unified, rector1996galen, DBLP:conf/amia/StearnsPSW01} largely focused on defining extensive ontologies for usage by biomedical experts.
BioCaster \cite{DBLP:journals/bioinformatics/CollierDKGCTNDKTST08} and PULS \cite{DBLP:conf/tsd/DuEKNTY11} explored utilizing rule-based methods for the news domain.
Early information extraction systems tackled predicting influenza trends from social media \cite{signorini2011use, lamb-etal-2013-separating, paul2014twitter}.
More recently, IDO \cite{DBLP:journals/biomedsem/BabcockBCS21} and DO \cite{DBLP:journals/nar/SchrimlMSOMFBJB22} are two extensive ontologies for human diseases.
CIDO \cite{DBLP:conf/icbo/He0OWLHHBLDAXHY20}, ExcavatorCovid \cite{min-etal-2021-excavatorcovid}, CACT \cite{DBLP:journals/jbi/LybargerOTY21} and COVIDKB \cite{zong-etal-2022-extracting, mendes-etal-2023-human} were developed specifically focused on COVID-19 events.
While most of these works are English-focused, some other works \cite{DBLP:journals/artmed/LejeuneBDL15, mutuvi-etal-2020-dataset, sahnoun-lejeune-2021-multilingual} support multilingual systems by using keyword-based and simple classification methods.
Overall, most of these systems are English-centric, disease-specific, not suitable for use for social media and utilize very rudimentary models.
In our work, utilizing disease-agnostic annotations and powerful multilingual models, we develop models that can detect events for any disease mentioned in any language.

\section{Conclusion and Future Work}

In our work, we pioneer the creation of the first multilingual Event Extraction (EE) framework for application in epidemic prediction and preparedness.
To this end, we create a multilingual EE benchmarking dataset \dataName{} comprising 5K tweets spanning four languages and four diseases.
To realistically deploy our models, we develop zero-shot cross-lingual cross-disease models and demonstrate their capability to extract events for 65 languages spanning 117 countries.
We prove the effectiveness of our model by providing early epidemic warnings for COVID-19 from Chinese Weibo posts in a zero-shot manner.
We also show evidence that our framework can be utilized as an information aggregation system aiding in misinformation detection and public attention monitoring.
In conclusion, we provide a strong utility of multilingual EE for global epidemic preparedness.

% \section*{Acknowledgements}

% Jia-Chen Gu, Di Wu, Po-Nien Kung, Alex Mac, Syed Shahriar, Rebecca (+2 other annotators), Mohsen Fayaaz, Rohan (if he reviews)

\section*{Acknowledgements}

We express our gratitude to Anh Mac, Syed Shahriar, Di Wu, Po-Nien Kung, Rohan Wadhawan, and Haw-Shiuan Chang for their valuable time, reviews of our work, and constructive feedback.
We thank the anonymous reviewers and
the area editors for their feedback.
This work was supported by NSF 2200274, 2106859, 2312501, DARPA HR00112290103/HR0011260656, NIH U54HG012517, U24DK097771, as well as Optum Labs. We thank them for their support.

\section*{Limitations}

% Our work focuses majorly on a single source of social media - Twitter.
% We haven't explored other social media platforms and how ED would work on those platforms in our work.
% We leave that for future work, but are optimistic that our models should be able to generalize across platforms.
% Secondly, our work mainly only focuses on ED as the primary task, while its sister task Event Argument Extraction (EAE) is not explored.
% We hope to extend our work for EAE as part of our future work.
% Finally, we would like to show the generalization of our models on a vast range of diseases.
% However owing to budget constraints and the lack of publically available Twitter data for other diseases, we couldn't perform such a study.
% However, we believe showing results on three diseases lays the foundation for generalizability of our model.

We benchmark our framework for four languages, but it is possible that it would be poor in performance for many others.
Owing to the lack of annotated data, it is difficult to conduct a holistic multilingual evaluation of our framework.
Our experiments on global epidemic prediction and information aggregation have been done on a single day of social media posts.
Furthermore, owing to the expensive cost of procuring massive social media data, it is infeasible run our framework across languages for a longer duration of time.
Finally, our major experiments are based on four diseases.
We would like to expand this further, but owing to budget constraints, we restrict ourselves to four diseases only in this work.

\section*{Ethical Considerations}

% One strong assumption in our work is the availability of internet and social media for discussions about epidemics.
% Since not everyone has equal access to these platforms, our dataset, models, and results do not represent the whole world uniformly.
% Thus, our work can be biased and should be considered with other sources for better representation.

% Our dataset \dataName{} is based on actual tweets posted by people all over the world.
% We attempted our best to anonymize any kind of private information in the tweets, but we can never be completely thorough, and there might be some private information embedded still in our dataset.
% Furthermore, these tweets were sentimental and may possess stark emotional, racial, and political viewpoints and biases.
% We do not attempt to clean any of such extreme data in our work (as our focus was on ED only) and these biases should be considered if being used for other applications.

% Since our ED models are trained on \dataName{}, they may possess some of the dataset-based social biases.
% Since we don't focus on bias mitigation, these models should be used with due consideration.

% Lastly, we do not claim that our models can be used off-the-shelf for epidemic prediction as it hasn't been thoroughly tested and can have false positives and negatives too.
% Furthermore, our results are shown on academic datasets and do not utilize all possible Twitter data.
% We majorly throw light to show these model capabilities and motivate future work in this direction.
% The usage of these systems for practical purposes should be appropriately considered.

Our framework extracts signals from social media wide range of languages to provide epidemic information.
However, the internet and access and usage of social media are disparate across the globe - leading to biased representation.
This aspect should be considered when utilizing our framework for inferring for low-resource languages or under-represented communities.

Since our work utilizes actual tweets, there could be some private information that could not be completely anonymized in our pre-processing.
These tweets may also possess stark emotional, racial, and political viewpoints and biases.
Our work doesn't focus on bias mitigation and our models may possess such a bias.
Due consideration should be taken when using our data or models.

Finally, despite our best efforts, our framework is far from being ideal and usable in real-life scenarios.
Our framework can output false positives frequently.
The goal of our work is to provide a strong prototype and encourage research in this direction.
Usage of our models/framework for practical use cases should be appropriately considered.

Note: We utilize ChatGPT in writing the paper better and correctly grammatical mistakes.

\bibliography{anthology,custom}

\begin{thebibliography}{66}
\providecommand{\natexlab}[1]{#1}

\bibitem[{Ahn(2006)}]{ahn-2006-stages}
David Ahn. 2006.
\newblock \href {https://aclanthology.org/W06-0901} {The stages of event extraction}.
\newblock In \emph{Proceedings of the Workshop on Annotating and Reasoning about Time and Events}, pages 1--8, Sydney, Australia. Association for Computational Linguistics.

\bibitem[{Babcock et~al.(2021)Babcock, Beverley, Cowell, and Smith}]{DBLP:journals/biomedsem/BabcockBCS21}
Shane Babcock, John Beverley, Lindsay~G. Cowell, and Barry Smith. 2021.
\newblock \href {https://doi.org/10.1186/s13326-021-00245-1} {The infectious disease ontology in the age of {COVID-19}}.
\newblock \emph{J. Biomed. Semant.}, 12(1):13.

\bibitem[{Bansal et~al.(2024)Bansal, Kung, Brantingham, Chang, and Peng}]{DBLP:journals/corr/abs-2404-04763}
Hritik Bansal, Po{-}Nien Kung, P.~Jeffrey Brantingham, Kai{-}Wei Chang, and Nanyun Peng. 2024.
\newblock \href {https://doi.org/10.48550/ARXIV.2404.04763} {Genearl: {A} training-free generative framework for multimodal event argument role labeling}.
\newblock \emph{CoRR}, abs/2404.04763.

\bibitem[{Brown et~al.(2020)Brown, Mann, Ryder, Subbiah, Kaplan, Dhariwal, Neelakantan, Shyam, Sastry, Askell, Agarwal, Herbert{-}Voss, Krueger, Henighan, Child, Ramesh, Ziegler, Wu, Winter, Hesse, Chen, Sigler, Litwin, Gray, Chess, Clark, Berner, McCandlish, Radford, Sutskever, and Amodei}]{DBLP:journals/corr/abs-2005-14165}
Tom~B. Brown, Benjamin Mann, Nick Ryder, Melanie Subbiah, Jared Kaplan, Prafulla Dhariwal, Arvind Neelakantan, Pranav Shyam, Girish Sastry, Amanda Askell, Sandhini Agarwal, Ariel Herbert{-}Voss, Gretchen Krueger, Tom Henighan, Rewon Child, Aditya Ramesh, Daniel~M. Ziegler, Jeffrey Wu, Clemens Winter, Christopher Hesse, Mark Chen, Eric Sigler, Mateusz Litwin, Scott Gray, Benjamin Chess, Jack Clark, Christopher Berner, Sam McCandlish, Alec Radford, Ilya Sutskever, and Dario Amodei. 2020.
\newblock \href {https://arxiv.org/abs/2005.14165} {Language models are few-shot learners}.
\newblock \emph{CoRR}, abs/2005.14165.

\bibitem[{Cai et~al.(2024)Cai, Kung, Suvarna, Ma, Bansal, Chang, Brantingham, Wang, and Peng}]{DBLP:journals/corr/abs-2403-02586}
Zefan Cai, Po{-}Nien Kung, Ashima Suvarna, Mingyu~Derek Ma, Hritik Bansal, Baobao Chang, P.~Jeffrey Brantingham, Wei Wang, and Nanyun Peng. 2024.
\newblock \href {https://doi.org/10.48550/ARXIV.2403.02586} {Improving event definition following for zero-shot event detection}.
\newblock \emph{CoRR}, abs/2403.02586.

\bibitem[{Collier et~al.(2008)Collier, Doan, Kawazoe, Goodwin, Conway, Tateno, Ngo, Dien, Kawtrakul, Takeuchi, Shigematsu, and Taniguchi}]{DBLP:journals/bioinformatics/CollierDKGCTNDKTST08}
Nigel Collier, Son Doan, Ai~Kawazoe, Reiko~Matsuda Goodwin, Mike Conway, Yoshio Tateno, Hung~Quoc Ngo, Dinh Dien, Asanee Kawtrakul, Koichi Takeuchi, Mika Shigematsu, and Kiyosu Taniguchi. 2008.
\newblock \href {https://doi.org/10.1093/bioinformatics/btn534} {Biocaster: detecting public health rumors with a web-based text mining system}.
\newblock \emph{Bioinform.}, 24(24):2940--2941.

\bibitem[{Conneau et~al.(2020)Conneau, Khandelwal, Goyal, Chaudhary, Wenzek, Guzmán, Grave, Ott, Zettlemoyer, and Stoyanov}]{conneau2020xlmroberta}
Alexis Conneau, Kartikay Khandelwal, Naman Goyal, Vishrav Chaudhary, Guillaume Wenzek, Francisco Guzmán, Edouard Grave, Myle Ott, Luke Zettlemoyer, and Veselin Stoyanov. 2020.
\newblock \href {https://arxiv.org/abs/1911.02116} {Unsupervised cross-lingual representation learning at scale}.
\newblock \emph{Preprint}, arXiv:1911.02116.

\bibitem[{Deng et~al.(2019)Deng, Zhang, Kang, Zhang, Zhang, and Chen}]{DBLP:journals/corr/abs-1910-11621}
Shumin Deng, Ningyu Zhang, Jiaojian Kang, Yichi Zhang, Wei Zhang, and Huajun Chen. 2019.
\newblock \href {https://arxiv.org/abs/1910.11621} {Meta-learning with dynamic-memory-based prototypical network for few-shot event detection}.
\newblock \emph{CoRR}, abs/1910.11621.

\bibitem[{Devlin et~al.(2019)Devlin, Chang, Lee, and Toutanova}]{devlin-etal-2019-bert}
Jacob Devlin, Ming-Wei Chang, Kenton Lee, and Kristina Toutanova. 2019.
\newblock \href {https://doi.org/10.18653/v1/N19-1423} {{BERT}: Pre-training of deep bidirectional transformers for language understanding}.
\newblock In \emph{Proceedings of the 2019 Conference of the North {A}merican Chapter of the Association for Computational Linguistics: Human Language Technologies, Volume 1 (Long and Short Papers)}, pages 4171--4186, Minneapolis, Minnesota. Association for Computational Linguistics.

\bibitem[{Dias(2020)}]{Dias2020}
Guilherme Dias. 2020.
\newblock \href {https://doi.org/10.6084/m9.figshare.12301400.v1} {{Tweets dataset on Zika virus}}.

\bibitem[{Doddington et~al.(2004)Doddington, Mitchell, Przybocki, Ramshaw, Strassel, and Weischedel}]{doddington-etal-2004-automatic}
George Doddington, Alexis Mitchell, Mark Przybocki, Lance Ramshaw, Stephanie Strassel, and Ralph Weischedel. 2004.
\newblock \href {http://www.lrec-conf.org/proceedings/lrec2004/pdf/5.pdf} {The automatic content extraction ({ACE}) program {--} tasks, data, and evaluation}.
\newblock In \emph{Proceedings of the Fourth International Conference on Language Resources and Evaluation ({LREC}{'}04)}, Lisbon, Portugal. European Language Resources Association (ELRA).

\bibitem[{Du et~al.(2011)Du, von Etter, Kopotev, Novikov, Tarbeeva, and Yangarber}]{DBLP:conf/tsd/DuEKNTY11}
Mian Du, Peter von Etter, Mikhail Kopotev, Mikhail Novikov, Natalia Tarbeeva, and Roman Yangarber. 2011.
\newblock \href {https://doi.org/10.1007/978-3-642-23538-2\_48} {Building support tools for russian-language information extraction}.
\newblock In \emph{Text, Speech and Dialogue - 14th International Conference, {TSD} 2011, Pilsen, Czech Republic, September 1-5, 2011. Proceedings}, volume 6836 of \emph{Lecture Notes in Computer Science}, pages 380--387. Springer.

\bibitem[{Du and Cardie(2020)}]{du-cardie-2020-event}
Xinya Du and Claire Cardie. 2020.
\newblock \href {https://doi.org/10.18653/v1/2020.emnlp-main.49} {Event extraction by answering (almost) natural questions}.
\newblock In \emph{Proceedings of the 2020 Conference on Empirical Methods in Natural Language Processing (EMNLP)}, pages 671--683, Online. Association for Computational Linguistics.

\bibitem[{Fleiss(1971)}]{fleiss1971measuring}
Joseph~L Fleiss. 1971.
\newblock \href {https://europepmc.org/article/CTX/c6682} {Measuring nominal scale agreement among many raters.}
\newblock \emph{Psychological bulletin}, 76(5):378.

\bibitem[{Garg et~al.(2018{\natexlab{a}})Garg, Parekh, and Jyothi}]{garg-etal-2018-code}
Saurabh Garg, Tanmay Parekh, and Preethi Jyothi. 2018{\natexlab{a}}.
\newblock \href {https://doi.org/10.18653/v1/D18-1346} {Code-switched language models using dual {RNN}s and same-source pretraining}.
\newblock In \emph{Proceedings of the 2018 Conference on Empirical Methods in Natural Language Processing}, pages 3078--3083, Brussels, Belgium. Association for Computational Linguistics.

\bibitem[{Garg et~al.(2018{\natexlab{b}})Garg, Parekh, and Jyothi}]{DBLP:conf/interspeech/GargPJ18}
Saurabh Garg, Tanmay Parekh, and Preethi Jyothi. 2018{\natexlab{b}}.
\newblock \href {https://doi.org/10.21437/INTERSPEECH.2018-1343} {Dual language models for code switched speech recognition}.
\newblock In \emph{19th Annual Conference of the International Speech Communication Association, Interspeech 2018, Hyderabad, India, September 2-6, 2018}, pages 2598--2602. {ISCA}.

\bibitem[{Grishman and Sundheim(1996)}]{grishman-sundheim-1996-message}
Ralph Grishman and Beth Sundheim. 1996.
\newblock \href {https://aclanthology.org/C96-1079} {{M}essage {U}nderstanding {C}onference- 6: A brief history}.
\newblock In \emph{{COLING} 1996 Volume 1: The 16th International Conference on Computational Linguistics}.

\bibitem[{Guo et~al.(2021)Guo, Fang, Zhou, Zhang, Guo, bi~Zeng, Chen, Liu, and Lu}]{Guo2021ImprovingGF}
Shuhui Guo, Fan Fang, Tao Zhou, Wei Zhang, Qiang Guo, Rui bi~Zeng, Xiaohong Chen, Jianguo Liu, and Xin Lu. 2021.
\newblock \href {https://api.semanticscholar.org/CorpusID:235915214} {Improving google flu trends for covid-19 estimates using weibo posts}.
\newblock \emph{Data Science and Management}, 3:13 -- 21.

\bibitem[{He et~al.(2020)He, Yu, Ong, Wang, Liu, Huffman, Huang, Beverley, Lin, Duncan, Arabandi, Xie, Hur, Yang, Chen, Omenn, Athey, and Smith}]{DBLP:conf/icbo/He0OWLHHBLDAXHY20}
Yongqun He, Hong Yu, Edison Ong, Yang Wang, Yingtong Liu, Anthony Huffman, Hsin{-}Hui Huang, John Beverley, Asiyah~Yu Lin, William~D. Duncan, Sivaram Arabandi, Jiangan Xie, Junguk Hur, Xiaolin Yang, Luonan Chen, Gilbert~S. Omenn, Brian~D. Athey, and Barry Smith. 2020.
\newblock \href {https://ceur-ws.org/Vol-2807/paperE.pdf} {{CIDO:} the community-based coronavirus infectious disease ontology}.
\newblock In \emph{Proceedings of the 11th International Conference on Biomedical Ontologies {(ICBO)} joint with the 10th Workshop on Ontologies and Data in Life Sciences {(ODLS)} and part of the Bolzano Summer of Knowledge (BoSK 2020), Virtual conference hosted in Bolzano, Italy, September 17, 2020}, volume 2807 of \emph{{CEUR} Workshop Proceedings}, pages 1--10. CEUR-WS.org.

\bibitem[{Heymann et~al.(2001)Heymann, Rodier et~al.}]{heymann2001hot}
David~L Heymann, Gu{\'e}na{\"e}l~R Rodier, et~al. 2001.
\newblock \href {https://pubmed.ncbi.nlm.nih.gov/11871807/} {Hot spots in a wired world: Who surveillance of emerging and re-emerging infectious diseases}.
\newblock \emph{The Lancet infectious diseases}, 1(5):345--353.

\bibitem[{Hsu et~al.(2021)Hsu, Guo, Natarajan, Peng et~al.}]{hsu2021discourse}
I~Hsu, Xiao Guo, Premkumar Natarajan, Nanyun Peng, et~al. 2021.
\newblock Discourse-level relation extraction via graph pooling.
\newblock \emph{arXiv preprint arXiv:2101.00124}.

\bibitem[{Hsu et~al.(2022)Hsu, Huang, Boschee, Miller, Natarajan, Chang, and Peng}]{hsu-etal-2022-degree}
I-Hung Hsu, Kuan-Hao Huang, Elizabeth Boschee, Scott Miller, Prem Natarajan, Kai-Wei Chang, and Nanyun Peng. 2022.
\newblock \href {https://doi.org/10.18653/v1/2022.naacl-main.138} {{DEGREE}: A data-efficient generation-based event extraction model}.
\newblock In \emph{Proceedings of the 2022 Conference of the North American Chapter of the Association for Computational Linguistics: Human Language Technologies}, pages 1890--1908, Seattle, United States. Association for Computational Linguistics.

\bibitem[{Hsu et~al.(2023{\natexlab{a}})Hsu, Huang, Zhang, Cheng, Natarajan, Chang, and Peng}]{hsu-etal-2023-tagprime}
I-Hung Hsu, Kuan-Hao Huang, Shuning Zhang, Wenxin Cheng, Prem Natarajan, Kai-Wei Chang, and Nanyun Peng. 2023{\natexlab{a}}.
\newblock \href {https://doi.org/10.18653/v1/2023.acl-long.723} {{TAGPRIME}: A unified framework for relational structure extraction}.
\newblock In \emph{Proceedings of the 61st Annual Meeting of the Association for Computational Linguistics (Volume 1: Long Papers)}, pages 12917--12932, Toronto, Canada. Association for Computational Linguistics.

\bibitem[{Hsu et~al.(2023{\natexlab{b}})Hsu, Xie, Huang, Natarajan, and Peng}]{hsu-etal-2023-ampere}
I-Hung Hsu, Zhiyu Xie, Kuan-Hao Huang, Prem Natarajan, and Nanyun Peng. 2023{\natexlab{b}}.
\newblock \href {https://doi.org/10.18653/v1/2023.acl-long.615} {{AMPERE}: {AMR}-aware prefix for generation-based event argument extraction model}.
\newblock In \emph{Proceedings of the 61st Annual Meeting of the Association for Computational Linguistics (Volume 1: Long Papers)}, pages 10976--10993, Toronto, Canada. Association for Computational Linguistics.

\bibitem[{Hsu et~al.(2024)Hsu, Xue, Pochhi, Bansal, Natarajan, Srinivasa, and Peng}]{hsu-etal-2024-argument}
I-Hung Hsu, Zihan Xue, Nilay Pochhi, Sahil Bansal, Prem Natarajan, Jayanth Srinivasa, and Nanyun Peng. 2024.
\newblock \href {https://doi.org/10.18653/v1/2024.findings-acl.758} {Argument-aware approach to event linking}.
\newblock In \emph{Findings of the Association for Computational Linguistics ACL 2024}, pages 12769--12781, Bangkok, Thailand and virtual meeting. Association for Computational Linguistics.

\bibitem[{Hu et~al.(2020)Hu, Huang, Chen, and Mao}]{hu-etal-2020-weibo}
Yong Hu, Heyan Huang, Anfan Chen, and Xian-Ling Mao. 2020.
\newblock \href {https://doi.org/10.18653/v1/2020.nlpcovid19-2.34} {{W}eibo-{COV}: A large-scale {COVID}-19 social media dataset from {W}eibo}.
\newblock In \emph{Proceedings of the 1st Workshop on {NLP} for {COVID}-19 (Part 2) at {EMNLP} 2020}, Online. Association for Computational Linguistics.

\bibitem[{Huang et~al.(2022)Huang, Hsu, Natarajan, Chang, and Peng}]{huang-etal-2022-multilingual-generative}
Kuan-Hao Huang, I-Hung Hsu, Prem Natarajan, Kai-Wei Chang, and Nanyun Peng. 2022.
\newblock \href {https://doi.org/10.18653/v1/2022.acl-long.317} {Multilingual generative language models for zero-shot cross-lingual event argument extraction}.
\newblock In \emph{Proceedings of the 60th Annual Meeting of the Association for Computational Linguistics (Volume 1: Long Papers)}, pages 4633--4646, Dublin, Ireland. Association for Computational Linguistics.

\bibitem[{Huang et~al.(2024)Huang, Hsu, Parekh, Xie, Zhang, Natarajan, Chang, Peng, and Ji}]{huang2024textee}
Kuan-Hao Huang, I-Hung Hsu, Tanmay Parekh, Zhiyu Xie, Zixuan Zhang, Premkumar Natarajan, Kai-Wei Chang, Nanyun Peng, and Heng Ji. 2024.
\newblock \href {https://arxiv.org/abs/2311.09562} {Textee: Benchmark, reevaluation, reflections, and future challenges in event extraction}.
\newblock \emph{Preprint}, arXiv:2311.09562.

\bibitem[{Lamb et~al.(2013)Lamb, Paul, and Dredze}]{lamb-etal-2013-separating}
Alex Lamb, Michael~J. Paul, and Mark Dredze. 2013.
\newblock \href {https://aclanthology.org/N13-1097} {Separating fact from fear: Tracking flu infections on {T}witter}.
\newblock In \emph{Proceedings of the 2013 Conference of the North {A}merican Chapter of the Association for Computational Linguistics: Human Language Technologies}, pages 789--795, Atlanta, Georgia. Association for Computational Linguistics.

\bibitem[{Lejeune et~al.(2015)Lejeune, Brixtel, Doucet, and Lucas}]{DBLP:journals/artmed/LejeuneBDL15}
Ga{\"{e}}l Lejeune, Romain Brixtel, Antoine Doucet, and Nadine Lucas. 2015.
\newblock \href {https://doi.org/10.1016/j.artmed.2015.06.005} {Multilingual event extraction for epidemic detection}.
\newblock \emph{Artif. Intell. Medicine}, 65(2):131--143.

\bibitem[{Li et~al.(2020)Li, Zareian, Zeng, Whitehead, Lu, Ji, and Chang}]{li-etal-2020-cross}
Manling Li, Alireza Zareian, Qi~Zeng, Spencer Whitehead, Di~Lu, Heng Ji, and Shih-Fu Chang. 2020.
\newblock \href {https://doi.org/10.18653/v1/2020.acl-main.230} {Cross-media structured common space for multimedia event extraction}.
\newblock In \emph{Proceedings of the 58th Annual Meeting of the Association for Computational Linguistics}, pages 2557--2568, Online. Association for Computational Linguistics.

\bibitem[{Li et~al.(2021)Li, Ji, and Han}]{li-etal-2021-document}
Sha Li, Heng Ji, and Jiawei Han. 2021.
\newblock \href {https://doi.org/10.18653/v1/2021.naacl-main.69} {Document-level event argument extraction by conditional generation}.
\newblock In \emph{Proceedings of the 2021 Conference of the North American Chapter of the Association for Computational Linguistics: Human Language Technologies}, pages 894--908, Online. Association for Computational Linguistics.

\bibitem[{Lin et~al.(2020{\natexlab{a}})Lin, Ji, Huang, and Wu}]{lin-etal-2020-joint}
Ying Lin, Heng Ji, Fei Huang, and Lingfei Wu. 2020{\natexlab{a}}.
\newblock \href {https://doi.org/10.18653/v1/2020.acl-main.713} {A joint neural model for information extraction with global features}.
\newblock In \emph{Proceedings of the 58th Annual Meeting of the Association for Computational Linguistics}, pages 7999--8009, Online. Association for Computational Linguistics.

\bibitem[{Lin et~al.(2020{\natexlab{b}})Lin, Ji, Huang, and Wu}]{lin-etal-2020-oneie}
Ying Lin, Heng Ji, Fei Huang, and Lingfei Wu. 2020{\natexlab{b}}.
\newblock \href {https://doi.org/10.18653/v1/2020.acl-main.713} {A joint neural model for information extraction with global features}.
\newblock In \emph{Proceedings of the 58th Annual Meeting of the Association for Computational Linguistics}, pages 7999--8009, Online. Association for Computational Linguistics.

\bibitem[{Lindberg et~al.(1993)Lindberg, Humphreys, and McCray}]{lindberg1993unified}
D.~A. Lindberg, B.~L. Humphreys, and A.~T. McCray. 1993.
\newblock \href {https://doi.org/10.1055/s-0038-1634945} {The unified medical language system.}
\newblock \emph{Methods of information in medicine}, 32(4):281–--291.

\bibitem[{Liu and Fu(2022)}]{liu2022study}
Lu~Liu and Yifei Fu. 2022.
\newblock Study on the mechanism of public attention to a major event: The outbreak of covid-19 in china.
\newblock \emph{Sustainable Cities and Society}, 81:103811.

\bibitem[{Lybarger et~al.(2021)Lybarger, Ostendorf, Thompson, and Yetisgen}]{DBLP:journals/jbi/LybargerOTY21}
Kevin Lybarger, Mari Ostendorf, Matthew Thompson, and Meliha Yetisgen. 2021.
\newblock \href {https://doi.org/10.1016/j.jbi.2021.103761} {Extracting {COVID-19} diagnoses and symptoms from clinical text: {A} new annotated corpus and neural event extraction framework}.
\newblock \emph{J. Biomed. Informatics}, 117:103761.

\bibitem[{Mendes et~al.(2023)Mendes, Chen, Xu, and Ritter}]{mendes-etal-2023-human}
Ethan Mendes, Yang Chen, Wei Xu, and Alan Ritter. 2023.
\newblock \href {https://doi.org/10.18653/v1/2023.acl-long.881} {Human-in-the-loop evaluation for early misinformation detection: A case study of {COVID}-19 treatments}.
\newblock In \emph{Proceedings of the 61st Annual Meeting of the Association for Computational Linguistics (Volume 1: Long Papers)}, pages 15817--15835, Toronto, Canada. Association for Computational Linguistics.

\bibitem[{Min et~al.(2021{\natexlab{a}})Min, Rozonoyer, Qiu, Zamanian, and MacBride}]{min2021excavatorcovid}
Bonan Min, Benjamin Rozonoyer, Haoling Qiu, Alexander Zamanian, and Jessica MacBride. 2021{\natexlab{a}}.
\newblock \href {https://arxiv.org/abs/2105.01819} {Excavatorcovid: Extracting events and relations from text corpora for temporal and causal analysis for covid-19}.
\newblock \emph{Preprint}, arXiv:2105.01819.

\bibitem[{Min et~al.(2021{\natexlab{b}})Min, Rozonoyer, Qiu, Zamanian, Xue, and MacBride}]{min-etal-2021-excavatorcovid}
Bonan Min, Benjamin Rozonoyer, Haoling Qiu, Alexander Zamanian, Nianwen Xue, and Jessica MacBride. 2021{\natexlab{b}}.
\newblock \href {https://doi.org/10.18653/v1/2021.emnlp-demo.8} {{E}xcavator{C}ovid: Extracting events and relations from text corpora for temporal and causal analysis for {COVID}-19}.
\newblock In \emph{Proceedings of the 2021 Conference on Empirical Methods in Natural Language Processing: System Demonstrations}, pages 63--71, Online and Punta Cana, Dominican Republic. Association for Computational Linguistics.

\bibitem[{Mutuvi et~al.(2020)Mutuvi, Doucet, Lejeune, and Odeo}]{mutuvi-etal-2020-dataset}
Stephen Mutuvi, Antoine Doucet, Ga{\"e}l Lejeune, and Moses Odeo. 2020.
\newblock \href {https://aclanthology.org/2020.lrec-1.509} {A dataset for multi-lingual epidemiological event extraction}.
\newblock In \emph{Proceedings of the Twelfth Language Resources and Evaluation Conference}, pages 4139--4144, Marseille, France. European Language Resources Association.

\bibitem[{Parekh et~al.(2020)Parekh, Ahn, Tsvetkov, and Black}]{parekh-etal-2020-understanding}
Tanmay Parekh, Emily Ahn, Yulia Tsvetkov, and Alan~W Black. 2020.
\newblock \href {https://doi.org/10.18653/v1/2020.conll-1.46} {Understanding linguistic accommodation in code-switched human-machine dialogues}.
\newblock In \emph{Proceedings of the 24th Conference on Computational Natural Language Learning}, pages 565--577, Online. Association for Computational Linguistics.

\bibitem[{Parekh et~al.(2023{\natexlab{a}})Parekh, Hsu, Huang, Chang, and Peng}]{clap}
Tanmay Parekh, I{-}Hung Hsu, Kuan{-}Hao Huang, Kai{-}Wei Chang, and Nanyun Peng. 2023{\natexlab{a}}.
\newblock \href {https://doi.org/10.48550/ARXIV.2309.08943} {Contextual label projection for cross-lingual structure extraction}.
\newblock \emph{CoRR}, abs/2309.08943.

\bibitem[{Parekh et~al.(2023{\natexlab{b}})Parekh, Hsu, Huang, Chang, and Peng}]{parekh-etal-2023-geneva}
Tanmay Parekh, I-Hung Hsu, Kuan-Hao Huang, Kai-Wei Chang, and Nanyun Peng. 2023{\natexlab{b}}.
\newblock \href {https://doi.org/10.18653/v1/2023.acl-long.203} {{GENEVA}: Benchmarking generalizability for event argument extraction with hundreds of event types and argument roles}.
\newblock In \emph{Proceedings of the 61st Annual Meeting of the Association for Computational Linguistics (Volume 1: Long Papers)}, pages 3664--3686, Toronto, Canada. Association for Computational Linguistics.

\bibitem[{Parekh et~al.(2024)Parekh, Mac, Yu, Dong, Shahriar, Liu, Yang, Huang, Wang, Peng, and Chang}]{speed}
Tanmay Parekh, Anh Mac, Jiarui Yu, Yuxuan Dong, Syed Shahriar, Bonnie Liu, Eric Yang, Kuan{-}Hao Huang, Wei Wang, Nanyun Peng, and Kai{-}Wei Chang. 2024.
\newblock \href {https://doi.org/10.48550/ARXIV.2404.01679} {Event detection from social media for epidemic prediction}.
\newblock \emph{CoRR}, abs/2404.01679.

\bibitem[{Paul et~al.(2014)Paul, Dredze, and Broniatowski}]{paul2014twitter}
Michael~J Paul, Mark Dredze, and David Broniatowski. 2014.
\newblock \href {https://pubmed.ncbi.nlm.nih.gov/25642377/} {Twitter improves influenza forecasting}.
\newblock \emph{PLoS currents}, 6.

\bibitem[{Pota et~al.(2021)Pota, Ventura, Fujita, and Esposito}]{DBLP:journals/eswa/PotaVFE21}
Marco Pota, Mirko Ventura, Hamido Fujita, and Massimo Esposito. 2021.
\newblock \href {https://doi.org/10.1016/J.ESWA.2021.115119} {Multilingual evaluation of pre-processing for bert-based sentiment analysis of tweets}.
\newblock \emph{Expert Syst. Appl.}, 181:115119.

\bibitem[{Pouran Ben~Veyseh et~al.(2022)Pouran Ben~Veyseh, Ebrahimi, Dernoncourt, and Nguyen}]{pouran-ben-veyseh-etal-2022-mee}
Amir Pouran Ben~Veyseh, Javid Ebrahimi, Franck Dernoncourt, and Thien Nguyen. 2022.
\newblock \href {https://doi.org/10.18653/v1/2022.emnlp-main.652} {{MEE}: A novel multilingual event extraction dataset}.
\newblock In \emph{Proceedings of the 2022 Conference on Empirical Methods in Natural Language Processing}, pages 9603--9613, Abu Dhabi, United Arab Emirates. Association for Computational Linguistics.

\bibitem[{Pyysalo et~al.(2012)Pyysalo, Ohta, Miwa, Cho, Tsujii, and Ananiadou}]{DBLP:journals/bioinformatics/PyysaloOMCTA12}
Sampo Pyysalo, Tomoko Ohta, Makoto Miwa, Han{-}Cheol Cho, Junichi Tsujii, and Sophia Ananiadou. 2012.
\newblock \href {https://doi.org/10.1093/BIOINFORMATICS/BTS407} {Event extraction across multiple levels of biological organization}.
\newblock \emph{Bioinform.}, 28(18):575--581.

\bibitem[{Rector et~al.(1996)Rector, Rogers, and Pole}]{rector1996galen}
Alan~L Rector, Jeremy~E Rogers, and Pam Pole. 1996.
\newblock \href {https://ebooks.iospress.nl/pdf/doi/10.3233/978-1-60750-878-6-174} {The galen high level ontology}.
\newblock In \emph{Medical Informatics Europe’96}, pages 174--178. IOS Press.

\bibitem[{Reimers and Gurevych(2019)}]{reimers-2019-sentence-bert}
Nils Reimers and Iryna Gurevych. 2019.
\newblock \href {https://arxiv.org/abs/1908.10084} {Sentence-bert: Sentence embeddings using siamese bert-networks}.
\newblock In \emph{Proceedings of the 2019 Conference on Empirical Methods in Natural Language Processing}. Association for Computational Linguistics.

\bibitem[{Sahnoun and Lejeune(2021)}]{sahnoun-lejeune-2021-multilingual}
Sihem Sahnoun and Ga{\"e}l Lejeune. 2021.
\newblock \href {https://aclanthology.org/2021.ranlp-1.138} {Multilingual epidemic event extraction : From simple classification methods to open information extraction ({OIE}) and ontology}.
\newblock In \emph{Proceedings of the International Conference on Recent Advances in Natural Language Processing (RANLP 2021)}, pages 1227--1233, Held Online. INCOMA Ltd.

\bibitem[{Schriml et~al.(2022)Schriml, Munro, Schor, Olley, McCracken, Felix, Baron, Jackson, Bello, Bearer, Lichenstein, Bisordi, Campion, Giglio, and Greene}]{DBLP:journals/nar/SchrimlMSOMFBJB22}
Lynn~M. Schriml, James~B. Munro, Mike Schor, Dustin Olley, Carrie McCracken, Victor Felix, J.~Allen Baron, Rebecca~C. Jackson, Susan~M. Bello, Cynthia Bearer, Richard Lichenstein, Katharine Bisordi, Nicole Campion, Michelle~G. Giglio, and Carol Greene. 2022.
\newblock \href {https://doi.org/10.1093/nar/gkab1063} {The human disease ontology 2022 update}.
\newblock \emph{Nucleic Acids Res.}, 50({D1}):1255--1261.

\bibitem[{Signorini et~al.(2011)Signorini, Segre, and Polgreen}]{signorini2011use}
Alessio Signorini, Alberto~Maria Segre, and Philip~M Polgreen. 2011.
\newblock \href {https://pubmed.ncbi.nlm.nih.gov/21573238/} {The use of twitter to track levels of disease activity and public concern in the us during the influenza a h1n1 pandemic}.
\newblock \emph{PloS one}, 6(5):e19467.

\bibitem[{Song et~al.(2015)Song, Bies, Strassel, Riese, Mott, Ellis, Wright, Kulick, Ryant, and Ma}]{song-etal-2015-light}
Zhiyi Song, Ann Bies, Stephanie Strassel, Tom Riese, Justin Mott, Joe Ellis, Jonathan Wright, Seth Kulick, Neville Ryant, and Xiaoyi Ma. 2015.
\newblock \href {https://doi.org/10.3115/v1/W15-0812} {From light to rich {ERE}: Annotation of entities, relations, and events}.
\newblock In \emph{Proceedings of the 3rd Workshop on {EVENTS}: Definition, Detection, Coreference, and Representation}, pages 89--98, Denver, Colorado. Association for Computational Linguistics.

\bibitem[{Stearns et~al.(2001)Stearns, Price, Spackman, and Wang}]{DBLP:conf/amia/StearnsPSW01}
Michael~Q. Stearns, Colin Price, Kent~A. Spackman, and Amy~Y. Wang. 2001.
\newblock \href {https://knowledge.amia.org/amia-55142-a2001a-1.597057/t-001-1.599654/f-001-1.599655/a-133-1.599740/a-134-1.599737} {{SNOMED} clinical terms: overview of the development process and project status}.
\newblock In \emph{{AMIA} 2001, American Medical Informatics Association Annual Symposium, Washington, DC, USA, November 3-7, 2001}. {AMIA}.

\bibitem[{Sundheim(1992)}]{sundheim-1992-overview}
Beth~M. Sundheim. 1992.
\newblock \href {https://aclanthology.org/M92-1001} {Overview of the fourth {M}essage {U}nderstanding {E}valuation and {C}onference}.
\newblock In \emph{{F}ourth {M}essage {U}nderstanding {C}onference ({MUC}-4): Proceedings of a Conference Held in {M}c{L}ean, {V}irginia, {J}une 16-18, 1992}.

\bibitem[{Suvarna et~al.(2024)Suvarna, Liu, Parekh, Chang, and Peng}]{DBLP:journals/corr/abs-2408-01046}
Ashima Suvarna, Xiao Liu, Tanmay Parekh, Kai{-}Wei Chang, and Nanyun Peng. 2024.
\newblock \href {https://doi.org/10.48550/ARXIV.2408.01046} {{QUDSELECT:} selective decoding for questions under discussion parsing}.
\newblock \emph{CoRR}, abs/2408.01046.

\bibitem[{Tong et~al.(2022)Tong, Xu, Wang, Han, Cao, Zhu, Chen, Hou, and Li}]{tong-etal-2022-docee}
MeiHan Tong, Bin Xu, Shuai Wang, Meihuan Han, Yixin Cao, Jiangqi Zhu, Siyu Chen, Lei Hou, and Juanzi Li. 2022.
\newblock \href {https://doi.org/10.18653/v1/2022.naacl-main.291} {{D}oc{EE}: A large-scale and fine-grained benchmark for document-level event extraction}.
\newblock In \emph{Proceedings of the 2022 Conference of the North American Chapter of the Association for Computational Linguistics: Human Language Technologies}, pages 3970--3982, Seattle, United States. Association for Computational Linguistics.

\bibitem[{Touvron et~al.(2023)Touvron, Martin, Stone, Albert, Almahairi, Babaei, Bashlykov, Batra, Bhargava, Bhosale, Bikel, Blecher, Canton{-}Ferrer, Chen, Cucurull, Esiobu, Fernandes, Fu, Fu, Fuller, Gao, Goswami, Goyal, Hartshorn, Hosseini, Hou, Inan, Kardas, Kerkez, Khabsa, Kloumann, Korenev, Koura, Lachaux, Lavril, Lee, Liskovich, Lu, Mao, Martinet, Mihaylov, Mishra, Molybog, Nie, Poulton, Reizenstein, Rungta, Saladi, Schelten, Silva, Smith, Subramanian, Tan, Tang, Taylor, Williams, Kuan, Xu, Yan, Zarov, Zhang, Fan, Kambadur, Narang, Rodriguez, Stojnic, Edunov, and Scialom}]{DBLP:journals/corr/abs-2307-09288}
Hugo Touvron, Louis Martin, Kevin Stone, Peter Albert, Amjad Almahairi, Yasmine Babaei, Nikolay Bashlykov, Soumya Batra, Prajjwal Bhargava, Shruti Bhosale, Dan Bikel, Lukas Blecher, Cristian Canton{-}Ferrer, Moya Chen, Guillem Cucurull, David Esiobu, Jude Fernandes, Jeremy Fu, Wenyin Fu, Brian Fuller, Cynthia Gao, Vedanuj Goswami, Naman Goyal, Anthony Hartshorn, Saghar Hosseini, Rui Hou, Hakan Inan, Marcin Kardas, Viktor Kerkez, Madian Khabsa, Isabel Kloumann, Artem Korenev, Punit~Singh Koura, Marie{-}Anne Lachaux, Thibaut Lavril, Jenya Lee, Diana Liskovich, Yinghai Lu, Yuning Mao, Xavier Martinet, Todor Mihaylov, Pushkar Mishra, Igor Molybog, Yixin Nie, Andrew Poulton, Jeremy Reizenstein, Rashi Rungta, Kalyan Saladi, Alan Schelten, Ruan Silva, Eric~Michael Smith, Ranjan Subramanian, Xiaoqing~Ellen Tan, Binh Tang, Ross Taylor, Adina Williams, Jian~Xiang Kuan, Puxin Xu, Zheng Yan, Iliyan Zarov, Yuchen Zhang, Angela Fan, Melanie Kambadur, Sharan Narang, Aur{\'{e}}lien Rodriguez, Robert Stojnic, Sergey Edunov,
  and Thomas Scialom. 2023.
\newblock \href {https://doi.org/10.48550/ARXIV.2307.09288} {Llama 2: Open foundation and fine-tuned chat models}.
\newblock \emph{CoRR}, abs/2307.09288.

\bibitem[{Wadden et~al.(2019{\natexlab{a}})Wadden, Wennberg, Luan, and Hajishirzi}]{wadden-etal-2019-entity}
David Wadden, Ulme Wennberg, Yi~Luan, and Hannaneh Hajishirzi. 2019{\natexlab{a}}.
\newblock \href {https://doi.org/10.18653/v1/D19-1585} {Entity, relation, and event extraction with contextualized span representations}.
\newblock In \emph{Proceedings of the 2019 Conference on Empirical Methods in Natural Language Processing and the 9th International Joint Conference on Natural Language Processing (EMNLP-IJCNLP)}, pages 5784--5789, Hong Kong, China. Association for Computational Linguistics.

\bibitem[{Wadden et~al.(2019{\natexlab{b}})Wadden, Wennberg, Luan, and Hajishirzi}]{Wadden2019dygiepp}
David Wadden, Ulme Wennberg, Yi~Luan, and Hannaneh Hajishirzi. 2019{\natexlab{b}}.
\newblock \href {https://api.semanticscholar.org/CorpusID:202539496} {Entity, relation, and event extraction with contextualized span representations}.
\newblock \emph{ArXiv}, abs/1909.03546.

\bibitem[{Wang et~al.(2020)Wang, Wang, Han, Jiang, Han, Liu, Li, Li, Lin, and Zhou}]{wang-etal-2020-maven}
Xiaozhi Wang, Ziqi Wang, Xu~Han, Wangyi Jiang, Rong Han, Zhiyuan Liu, Juanzi Li, Peng Li, Yankai Lin, and Jie Zhou. 2020.
\newblock \href {https://doi.org/10.18653/v1/2020.emnlp-main.129} {{MAVEN}: {A} {M}assive {G}eneral {D}omain {E}vent {D}etection {D}ataset}.
\newblock In \emph{Proceedings of the 2020 Conference on Empirical Methods in Natural Language Processing (EMNLP)}, pages 1652--1671, Online. Association for Computational Linguistics.

\bibitem[{Wu et~al.(2024)Wu, Shen, and Chang}]{DBLP:journals/corr/abs-2407-00191}
Di~Wu, Xiaoxian Shen, and Kai{-}Wei Chang. 2024.
\newblock \href {https://doi.org/10.48550/ARXIV.2407.00191} {Metakp: On-demand keyphrase generation}.
\newblock \emph{CoRR}, abs/2407.00191.

\bibitem[{Xue et~al.(2021)Xue, Constant, Roberts, Kale, Al-Rfou, Siddhant, Barua, and Raffel}]{xue2021mt5}
Linting Xue, Noah Constant, Adam Roberts, Mihir Kale, Rami Al-Rfou, Aditya Siddhant, Aditya Barua, and Colin Raffel. 2021.
\newblock \href {https://arxiv.org/abs/2010.11934} {mt5: A massively multilingual pre-trained text-to-text transformer}.
\newblock \emph{Preprint}, arXiv:2010.11934.

\bibitem[{Zong et~al.(2022)Zong, Baheti, Xu, and Ritter}]{zong-etal-2022-extracting}
Shi Zong, Ashutosh Baheti, Wei Xu, and Alan Ritter. 2022.
\newblock \href {https://aclanthology.org/2022.coling-1.335} {Extracting a knowledge base of {COVID}-19 events from social media}.
\newblock In \emph{Proceedings of the 29th International Conference on Computational Linguistics}, pages 3810--3823, Gyeongju, Republic of Korea. International Committee on Computational Linguistics.

\end{thebibliography}
% \bibliography{custom}

\pagebreak
\clearpage

\appendix

\appendix

\section{Ontology Creation: Role Definitions}
\label{sec:appendix-ontology}
% \tanmay{@jiarui Add the complete ontology table with argument definitions and examples}

We provide our complete event ontology, including argument definitions along with corresponding examples in Table~\ref{tab:infect-ontology}-\ref{tab:death-ontology}. We underline the arguments corresponding to each role in the examples.
We note that our ontology can be further utilized for other tasks as well, like relation extraction \cite{hsu2021discourse} and event linking \cite{hsu-etal-2024-argument}.

\section{Dataset Filtering and Sampling}
\label{sec:appendix-data-processing}
% \tanmay{@sreya add additional details here}

While there are works which focus on Event Extraction from multimodal tweets \cite{DBLP:journals/corr/abs-2404-04763}, we restrict our work to text-based tweets only.
We associate each event with 5-10 seed tweets inspired by SPEED \cite{speed}.
Utilizing embedding-space similarity of query tweets and our seed tweet repository, we filter out tweets related to epidemic events.
For multilingual languages, we translate the English seed tweets into individual languages.
We modify and further correct the translations with the help of human experts.
We provide some seed tweets per language per event in Table~\ref{tab:seed-tweets}.
% \tanmay{@sreya, can you make this table?}
We utilize the sentence-transformer model \cite{reimers-2019-sentence-bert} for embedding the tweets.

Furthermore, we utilize the event-similarity to uniformly sample tweets based on events.
More specifically, we over-sample tweets from frequent events and under-sample for the non-frequent ones.
Such uniform sampling has proved elemental to more robust model training, as noted in \citet{parekh-etal-2023-geneva}.

% \hyosang{Some sentences in section C.1 are written in past tense where it should be in present tense}

\section{Annotation Guidelines and Details}
\label{sec:appendix-data-annotation}

\begin{figure*}[h]
    \centering
    \includegraphics[width=\linewidth]{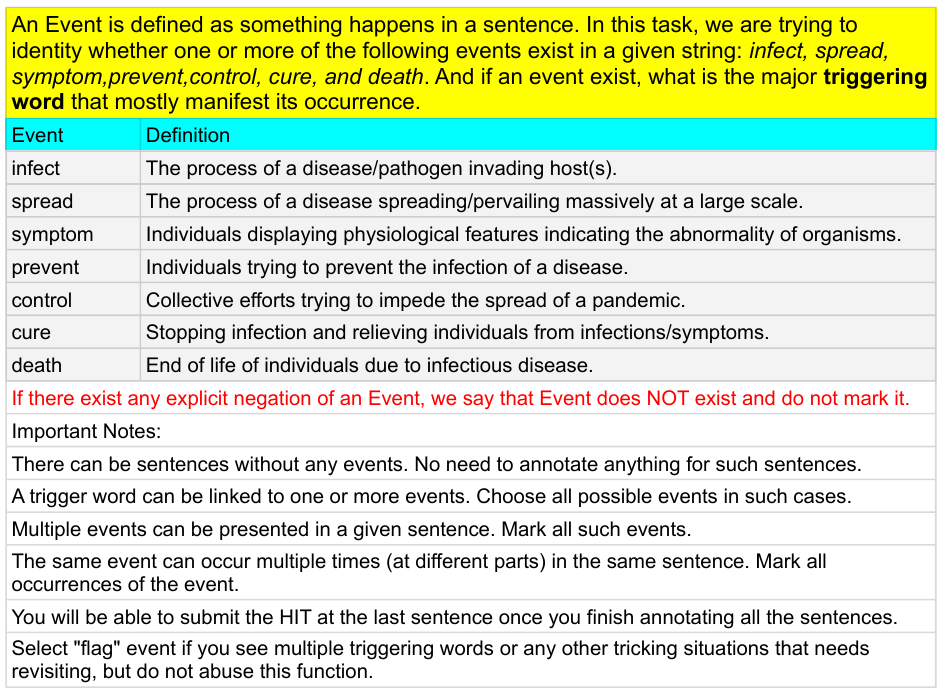}
    \caption{Guidelines for ED annotations for the \dataName{} dataset.}
    \label{fig:ed-guidelines}
\end{figure*}

We conduct two sets of annotations in our work and describe both in more detail here.
First, we conduct ED annotations for multilingual data in Japanese, Hindi, and Spanish.
We refer to ACE \cite{doddington-etal-2004-automatic} and SPEED \cite{speed} to chalk our guidelines.
We provide instructions and examples to the annotators in English while they are expected to annotate data in the respective target languages.
We provide the exact annotation guidelines in Figure~\ref{fig:ed-guidelines}.
% \tanmay{@jiarui add the guideline figure}

\begin{figure*}[t]
    \centering
    \includegraphics[width=1\linewidth]{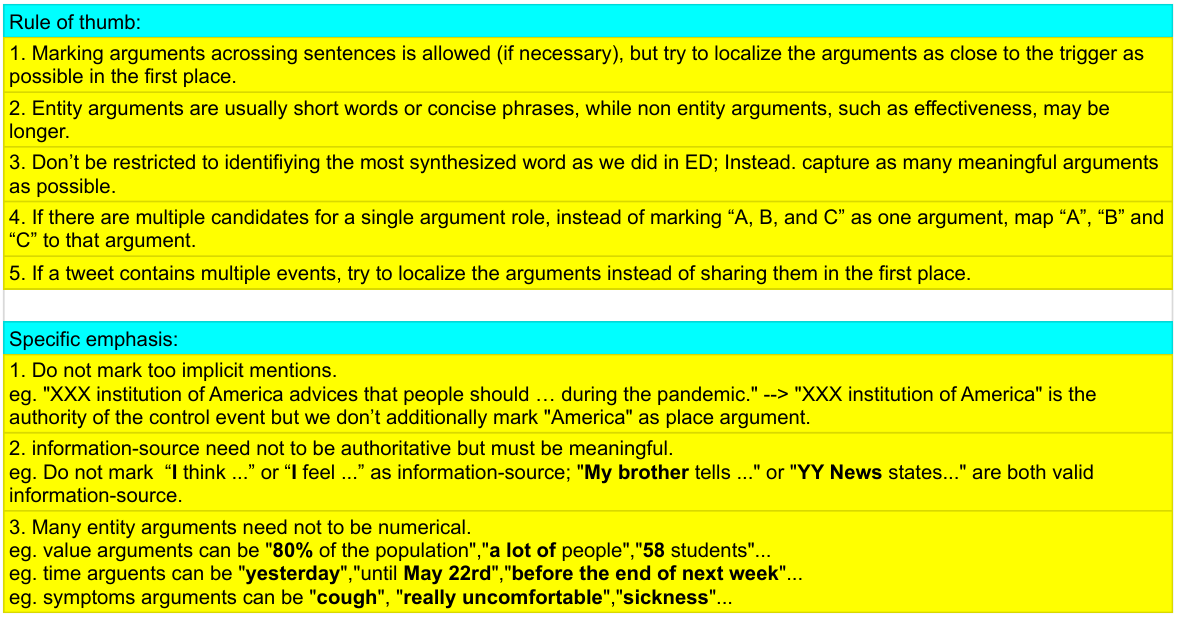}
    \caption{Guidelines for EAE annotation for the \dataName{} dataset.}
    \label{fig:eae-guidelines}
\end{figure*}

Next, we conduct EAE annotations for all the languages.
Inspired by ACE \cite{doddington-etal-2004-automatic} and GENEVA \cite{parekh-etal-2023-geneva}, we design our guidelines with special instructions.
We present the instructions in Figure~\ref{fig:eae-guidelines} along with simple argument definitions in Figure~\ref{fig:annotation-arg-examples}.

% Finally, for the multilingual verification process, we utilize partial guidelines from ED and EAE both while simplifying the task as the annotators are not experts.
% We use the simplified argument definitions, as shown in Figure~\ref{fig:annotation-arg-examples} along with the instructions (Figure~\ref{fig:verification-instruction}) and examples (Figure~\ref{fig:verification-examples}) for this annotation.

% \tanmay{Add screenshots of annotation guidelines and interface. Add some details about the task definitions for ED/EAE more clearly. Discuss the guidelines for multilingual data verification}

\begin{table}[t]
    \centering
    \small
    \begin{tabular}{|l|p{2.5cm}|p{2.5cm}|}
        \toprule
        \textbf{Language} & \textbf{Inconsistent rate} & \textbf{Verification acceptance rate} \\ \midrule
        Hindi    & 27.46\%                                 & 31.48\%                                           \\ \midrule
        Japanese & 17.52\%                                 & 81.73\%                                            \\ \midrule
        Spanish  & 19.01\%                                & 66.66\%                                            \\ \bottomrule
    \end{tabular}
    \caption{Inconsistencies identified (as percentage) and verifications accepted (as percentage) for the multilingual verifications. The inconsistent rate is the percentage of annotations with which the bilingual speakers did not agree, while the verification rate is the percentage of suggestions from the bilingual speakers that are accepted by our multilingual annotators.}
    \label{tab:annotator-verifier-agreement}
\end{table}

\subsection{Multilingual Data Verification}
\label{sec:appendix-multilingual-data-verification}

Due to the scarcity of multilingual annotators, we adopted a verification process different from the English data verification.
The entire verification procedure can be divided into four phases: qualification task, inter-annotator agreement (IAA) study, verification task, and correction task.

\begin{figure*}
    \centering
    \includegraphics[width=1\linewidth]{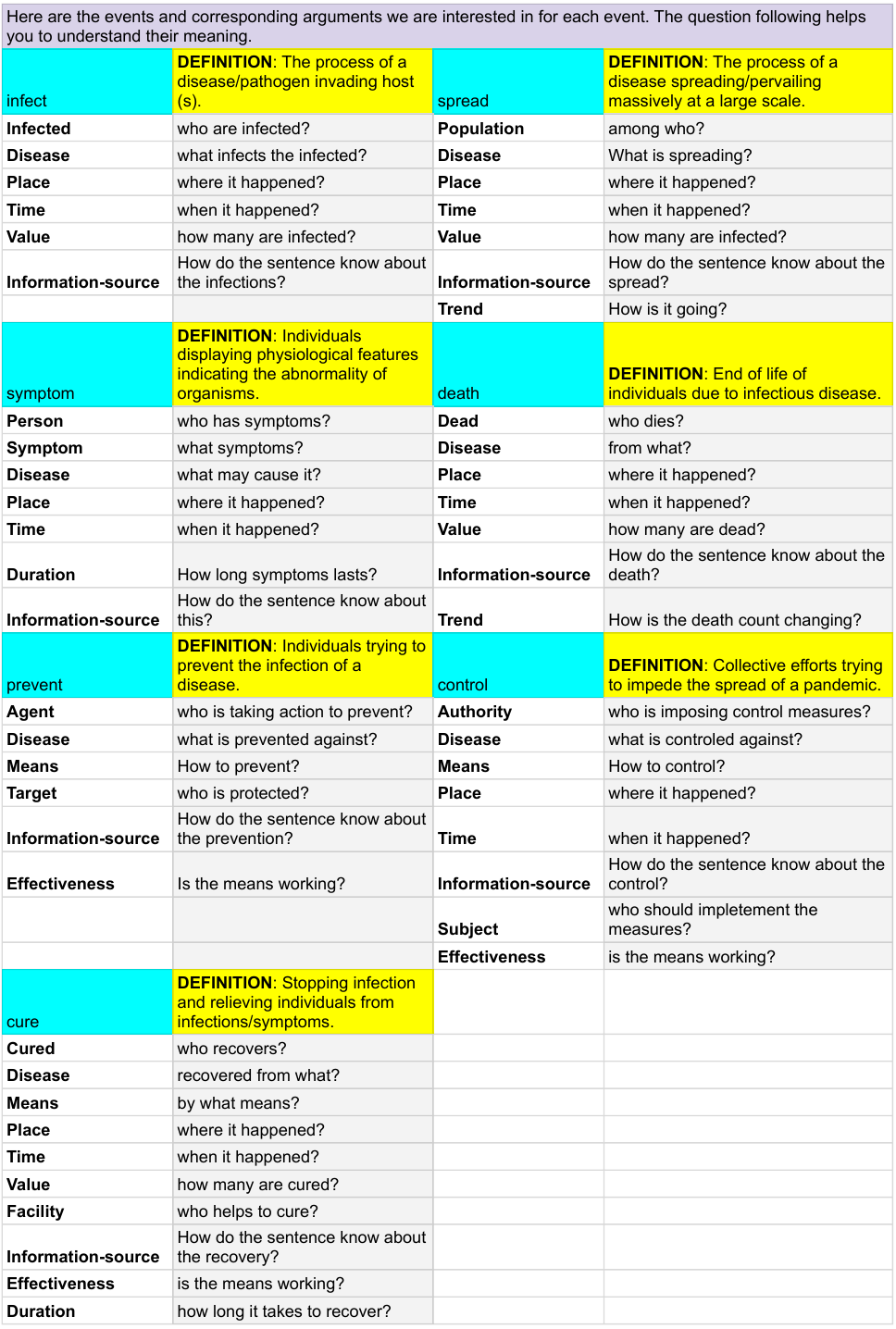}
    \caption{Argument definitions provided as part of the EAE annotation process.}
    \label{fig:annotation-arg-examples}
\end{figure*}

\begin{figure*}[t]
    \centering
    \includegraphics[width=1\linewidth]{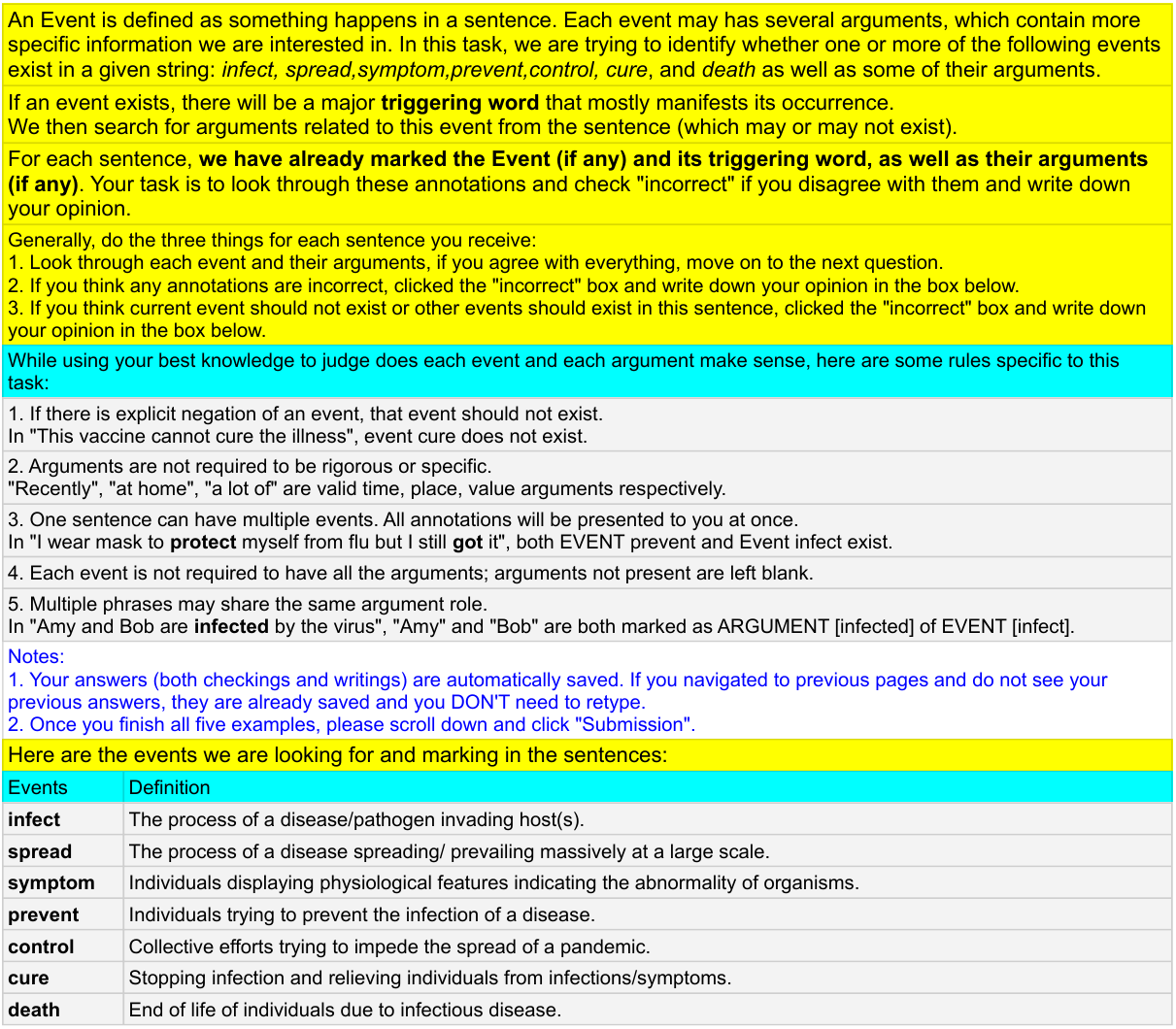}
    \caption{Intructions provided for the multilingual verification task.}
    \label{fig:verification-instruction}
\end{figure*}

\begin{figure*}
    \centering
    \includegraphics[width=1\linewidth]{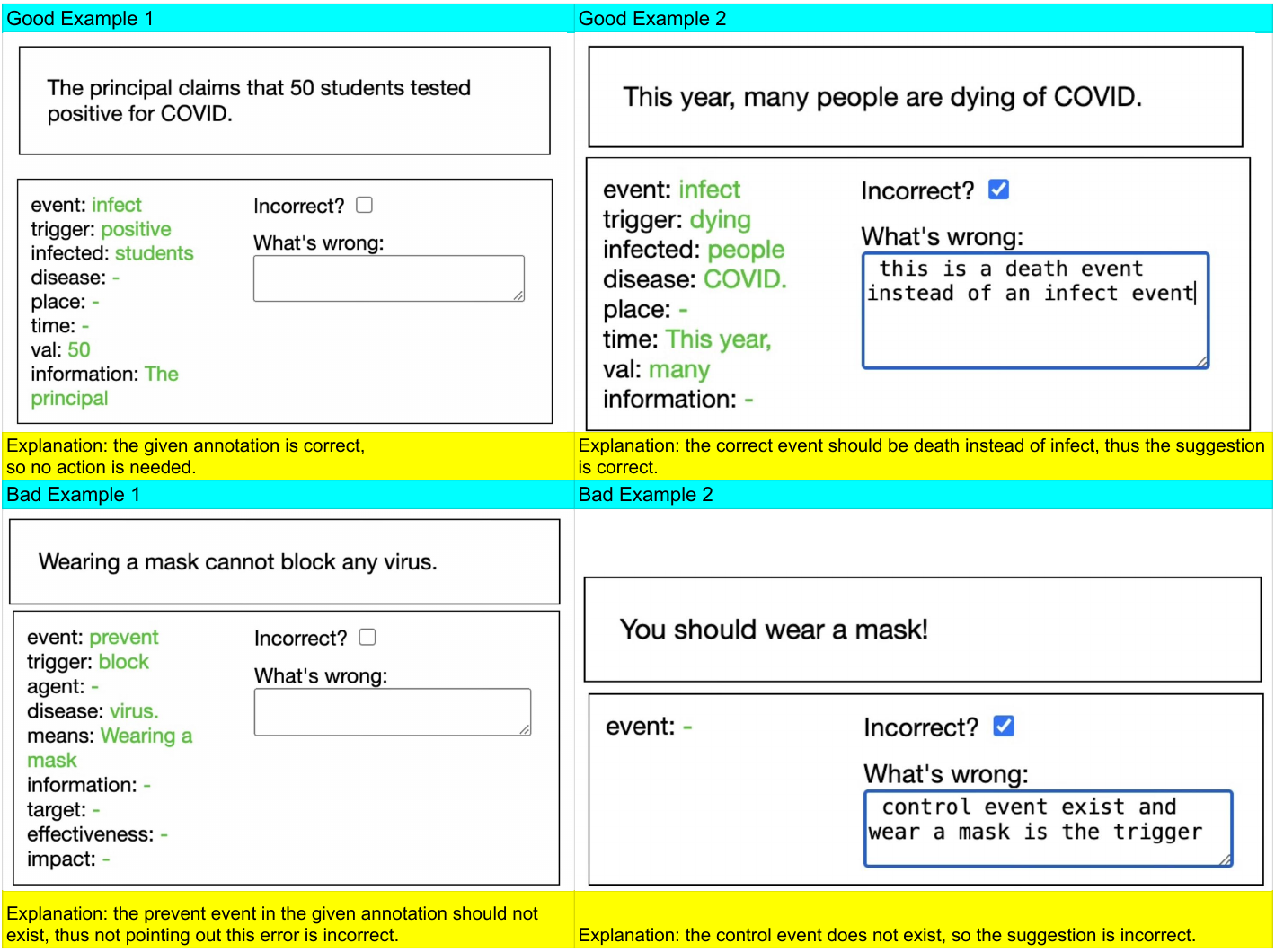}
    \caption{Illustrations provided for the multilingual verification task.}
    \label{fig:verification-examples}
\end{figure*}

We opt to choose bilingual speakers of English and Hindi/Japanese/Spanish as the verifiers.
Not that we do not consider code-switching \cite{garg-etal-2018-code, DBLP:conf/interspeech/GargPJ18} in our work, but bilingualism helps to ensure that the instructions are well understood by the verifiers.
To ensure verification quality, each bilingual speaker must pass a qualification test before entering the verification process.
They are provided with a guideline explaining their primary task introducing the essence ED/EAE annotation (Figure~\ref{fig:verification-instruction}), argument definitions (Figure~\ref{fig:annotation-arg-examples}) along with two pairs of positive and negative examples (Figure~\ref{fig:verification-examples}) - all in English.
Although not directly tasked with ED/EAE annotation, they must understand the standards of ED/EAE annotation to fairly judge the correctness of a given annotated example.
After reading the instructions, the bilingual speaker must correctly answer at least 4 out of 5 test questions to pass the qualification test.
Selected by our multilingual annotators, these test questions are in Japanese, Hindi, and Spanish, respectively, and are in the same format as the verification questions.
Failing the qualification task indicates an insufficient understanding of the verification task, thereby disqualifying the bilingual speaker from proceeding further.
We select one verifier for each language after filtering from this round.
Each of the verifiers was paid \$150 in total for 6 hours of their service at the rate of \$25/hr in line with other works \cite{parekh-etal-2020-understanding}.

Next, we asked the three qualified bilingual speakers to verify 40 English examples as part of an inter-annotator agreement (IAA) study to ensure an adequate agreement rate among them.
These IAA examples are in the same format as the actual verification questions.
They reached a final IAA score of 0.6 on the 40 English samples.

Next, the qualified bilingual speakers participate in the final verification process.
Along with the tweet text, they are shown all the events and arguments identified by our annotators.
If they agree with the current annotation, no action is needed; otherwise, they should check the “incorrect” box and provide their reasons (as shown in Figure~\ref{fig:verification-examples}).

Following the verification process, our multilingual annotators addressed the correction task: reviewing the comments and deciding on the final annotation.
We provide statistics about the total corrections suggested and accepted by the multilingual annotators for each language in Table~\ref{tab:annotator-verifier-agreement}.
% Below (table) are the agreement rates between the annotators and bilingual speakers.

% \tanmay{@jiarui add table here}

\section{Benchmarking Model: Implementation Details}
\label{sec:appendix-implementation-details}

% \tanmay{@hyosang add details about the different models here. You can take inspiration from the original SPEED paper.}

We use the EE benchmarking tool TextEE~\cite{huang2024textee} to conduct the benchmarking experiment of the models. We present details about each ED, EAE, and end-to-end model that we benchmark, along with the extensive set of hyperparameters and other implementation details.

\subsection{TagPrime}

TagPrime~\cite{hsu-etal-2023-tagprime} is a sequence tagging model with a word priming technique to convey more task-specific information. We run our experiments on the ED and EAE tasks of TagPrime on an NVIDIA RTX A6000 machine with support for 8 GPUs. The models are fine-tuned on XLM-RoBERTa-Large~\cite{conneau2020xlmroberta}.
We train this model separately for ED and EAE.
The major hyperparameters are listed in Table~\ref{tab:hyper-tagprime-ed} for the ED model and Table~\ref{tab:hyper-tagprime-eae} for the EAE model.

\begin{table}[h]
    \centering
    \small
    \begin{tabular}{lr}
        \toprule
        Pre-trained LM & XLM-RoBERTa-Large \\
        Training Batch Size & $16$ \\
        Eval Batch Size & $4$ \\
        Learning Rate & $0.001$ \\
        Weight Decay & $0.001$ \\
        Gradient Clipping & $5$ \\
        Training Epochs & $10$ \\
        Warmup Epochs & $5$ \\
        Max Sequence Length & $250$ \\
        Linear Layer Dropout & $0.2$ \\
        \bottomrule
    \end{tabular}
    \caption{Hyperparameter details for TagPrime ED model.}
    \label{tab:hyper-tagprime-ed}
\end{table}

\begin{table}[h]
    \centering
    \small
    \begin{tabular}{lr}
        \toprule
        Pre-trained LM & XLM-RoBERTa-Large \\
        Training Batch Size & $6$ \\
        Eval Batch Size & $12$ \\
        Learning Rate & $0.001$ \\
        Weight Decay & $0.001$ \\
        Gradient Clipping & $5$ \\
        Training Epochs & $90$ \\
        Warmup Epochs & $5$ \\
        Max Sequence Length & $250$ \\
        Linear Layer Dropout & $0.2$ \\
        \bottomrule
    \end{tabular}
    \caption{Hyperparameter details for TagPrime EAE model.}
    \label{tab:hyper-tagprime-eae}
\end{table}

\subsection{XGear}

XGear~\cite{huang-etal-2022-multilingual-generative} is a language-agnostic model that models EAE as a generation task. This model is similar to other generative models like DEGREE \cite{hsu-etal-2022-degree} and AMPERE \cite{hsu-etal-2023-ampere} but focuses on zero-shot cross-lingual transfer. We run our experiments on the EAE tasks of XGear on an NVIDIA RTX A6000 machine with support for 8 GPUs. The model is fine-tuned on mT5-Large~\cite{xue2021mt5}. The major hyperparameters for this model are listed in Table~\ref{tab:hyper-xgear}. To evaluate its end-to-end performance, we complement it with the TagPrime ED model.

\begin{table}[h]
    \centering
    \small
    \begin{tabular}{lr}
        \toprule
        Pre-trained LM & mT5-Large \\
        Training Batch Size & $6$ \\
        Eval Batch Size & $12$ \\
        Learning Rate & $0.00001$ \\
        Weight Decay & $0.00001$ \\
        Gradient Clipping & $5$ \\
        Training Epochs & $90$ \\
        Warmup Epochs & $5$ \\
        Max Sequence Length & $400$ \\
        % Linear Layer Dropout & $0.2$ \\
        \bottomrule
    \end{tabular}
    \caption{Hyperparameter details for XGear EAE model.}
    \label{tab:hyper-xgear}
\end{table}

\subsection{BERT-QA}

BERT-QA~\cite{du-cardie-2020-event} is a classification model utilizing label semantics via transforming the EE task into a question-answer task. We run our experiments on the EAE tasks of BERT-QA with support for 8 GPUs. The model is fine-tuned on XLM-RoBERTa-Large~\cite{conneau2020xlmroberta}. The major hyperparameters for this model are listed in Table~\ref{tab:hyper-bertqa-ed} for the ED model and Table~\ref{tab:hyper-bertqa-eae} for the EAE model.

\begin{table}[h]
    \centering
    \small
    \begin{tabular}{lr}
        \toprule
        Pre-trained LM & XLM-RoBERTa-Large \\
        Training Batch Size & $6$ \\
        Eval Batch Size & $12$ \\
        Learning Rate & $0.001$ \\
        Weight Decay & $0.001$ \\
        Gradient Clipping & $5$ \\
        Training Epochs & $30$ \\
        Warmup Epochs & $5$ \\
        Max Sequence Length & $250$ \\
        Linear Layer Dropout & $0.2$ \\
        \bottomrule
    \end{tabular}
    \caption{Hyperparameter details for BERT-QA ED model.}
    \label{tab:hyper-bertqa-ed}
\end{table}

\begin{table}[h]
    \centering
    \small
    \begin{tabular}{lr}
        \toprule
        Pre-trained LM & XLM-RoBERTa-Large \\
        Training Batch Size & $6$ \\
        Eval Batch Size & $12$ \\
        Learning Rate & $0.00001$ \\
        Weight Decay & $0.00001$ \\
        Gradient Clipping & $5$ \\
        Training Epochs & $90$ \\
        Warmup Epochs & $5$ \\
        Max Sequence Length & $400$ \\
        Linear Layer Dropout & $0.2$ \\
        \bottomrule
    \end{tabular}
    \caption{Hyperparameter details for BERT-QA EAE model.}
    \label{tab:hyper-bertqa-eae}
\end{table}

\subsection{DyGIE++}

DyGIE++~\cite{Wadden2019dygiepp} is an end-to-end model that simultaneously leverages span graph propagation for EE, entity recognition, and relation extraction tasks. We run our experiments on the end-to-end tasks of DyGIE++ on an NVIDIA RTX A6000 machine with support for 8 GPUs. The model is fine-tuned on XLM-RoBERTa-Large~\cite{conneau2020xlmroberta}. The major hyperparameters are listed in Table~\ref{tab:hyper-dygiepp}.

\begin{table}[h]
    \centering
    \small
    \begin{tabular}{lr}
        \toprule
        Pre-trained LM & XLM-RoBERTa-Large \\
        Training Batch Size & $6$ \\
        Eval Batch Size & $12$ \\
        Learning Rate & $0.001$ \\
        Weight Decay & $0.001$ \\
        Gradient Clipping & $5$ \\
        Training Epochs & $60$ \\
        Warmup Epochs & $5$ \\
        Max Sequence Length & $250$ \\
        Linear Layer Dropout & $0.4$ \\
        \bottomrule
    \end{tabular}
    \caption{Hyperparameter details for DyGIE++ end-to-end model.}
    \label{tab:hyper-dygiepp}
\end{table}

\subsection{OneIE}

OneIE~\cite{lin-etal-2020-oneie} is an end-to-end model that extracts a globally optimal information network from input sentences to capture interactions among entities, relations, and events. We run our experiments on the end-to-end tasks of OneIE on an NVIDIA RTX A6000 machine with support for 8 GPUs. The models are fine-tuned on XLM-RoBERTa-Large~\cite{conneau2020xlmroberta}. The major hyperparameters are listed in Table~\ref{tab:hyper-oneie}.

\begin{table}[h]
    \centering
    \small
    \begin{tabular}{lr}
        \toprule
        Pre-trained LM & XLM-RoBERTa-Large \\
        Training Batch Size & $6$ \\
        Eval Batch Size & $10$ \\
        Learning Rate & $0.001$ \\
        Weight Decay & $0.001$ \\
        Gradient Clipping & $5$ \\
        Training Epochs & $60$ \\
        Warmup Epochs & $5$ \\
        Max Sequence Length & $250$ \\
        Linear Layer Dropout & $0.4$ \\
        \bottomrule
    \end{tabular}
    \caption{Hyperparameter details for OneIE end-to-end model.}
    \label{tab:hyper-oneie}
\end{table}

\subsection{CLaP}

CLaP~\cite{clap} is a multilingual data-augmentation technique for structured prediction tasks utilizing constrained machine translation for label projection.
Specifically, we translate the English \dataName{} into other languages using CLaP.
We utilize five in-context examples for each language - Hindi, Japanese, and Spanish - with the original CLaP prompt using the Llama2-13B \cite{DBLP:journals/corr/abs-2307-09288} model.
We apply post-processing from \dataName{} to reduce the distribution difference between \dataName{} and the generated multilingual data.
We train a separate model for each language as it provided better results than joint training on all language data.

\subsection{DivED}

DivED \cite{DBLP:journals/corr/abs-2403-02586} is trained with the LLaMA-2-7B \cite{DBLP:journals/corr/abs-2307-09288} models on DivED and GENEVA \cite{parekh-etal-2023-geneva} dataset for zero-shot event detection. They utilize 200 + 90 event types from DivED and Geneva datasets respectively. Training is done using ten event definitions, ten samples, and ten negative samples per sample for each event type while incorporating the ontology information and three hard-negative samples.
We utilize their available trained model for our experiments.

\subsection{COVIDKB}

COVIDKB~\cite{zong-etal-2022-extracting} is a simple BERT-classification model trained on a multi-label classification objective on the COVIDKB Twitter corpus.
Since our ontology differs from their model, we train it as a binary classification model.
We run our experiments on the end-to-end tasks of this model on an NVIDIA RTX A6000 machine with 8 GPUs.
The model is fine-tuned on the multilingual BERT~\cite{devlin-etal-2019-bert}.
The major hyperparameters are listed in Table~\ref{tab:hyper-covidkb}.

\begin{table}[h]
    \centering
    \small
    \begin{tabular}{lr}
        \toprule
        Pre-trained LM & mBERT \\
        Training Batch Size & $64$ \\
        Learning Rate & $0.00002$ \\
        Training Epochs & $4$ \\
        Max Sequence Length & $128$ \\
        Number of Classes & $2$ \\
        \bottomrule
    \end{tabular}
    \caption{Hyperparameter details for COVIDKB binary classification model.}
    \label{tab:hyper-covidkb}
\end{table}

\begin{figure}[t]
    \centering
    \includegraphics[width=\columnwidth]{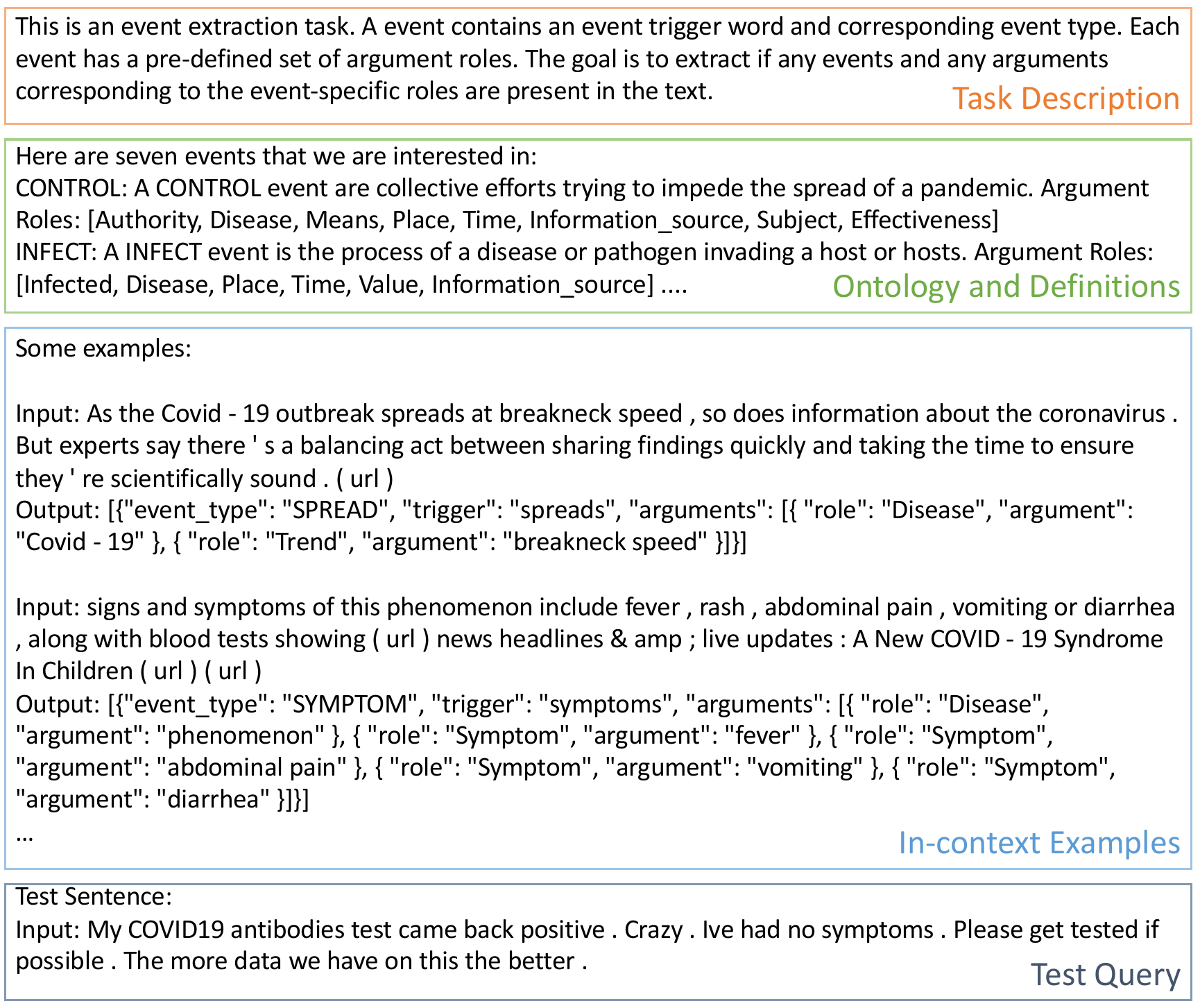}
    \caption{Illustration of the prompt used for GPT-3.5 model. It includes a task description, followed by ontology details of event types and their definitions. Next, we show some in-context examples for each event type and, finally, provide the test sentence.}
    \label{fig:ee-gpt-prompt}
\end{figure}

\subsection{Keyword}

This model curates a list of keywords specific to each event and predicts a trigger for a particular event if it matches one of the curated event keywords.
We utilize the base set of keywords from SPEED \cite{speed} for English.
We translate these English event-specific keywords for other languages.
Recent works have developed advanced keyword extraction techniques \cite{DBLP:journals/corr/abs-2407-00191}. However, we couldn't explore them in the scope of our work.

\begin{table}[t]
    \centering
    \small
    \setlength{\tabcolsep}{1.9pt}
    \begin{tabular}{l|c|c|c|c|c|c|c|c|c}
        \toprule
        \multirow{3}{*}{\textbf{Model}} & \multicolumn{3}{c|}{\textbf{COVID}} & \multicolumn{1}{c|}{\textbf{MPox}} & \multicolumn{4}{c|}{\textbf{Zika + Dengue}} & \multicolumn{1}{c}{\textbf{Avg}} \\
        & \multicolumn{1}{c}{hi} & \multicolumn{1}{c}{jp} & \multicolumn{1}{c|}{es} & \multicolumn{1}{c|}{en} & \multicolumn{1}{c}{en} & \multicolumn{1}{c}{hi} & \multicolumn{1}{c}{jp} & \multicolumn{1}{c|}{es} \\ 
        % \cmidrule(r){2-10}
        % & \textbf{TC} & \textbf{TC} & \textbf{TC} & \textbf{TC} & \textbf{TC} & \textbf{TC} & \textbf{TC} & \textbf{TC} & \textbf{TC} \\
        \midrule
        \multicolumn{10}{c}{\small \textsc{Baseline Models}} \\
        \midrule
        ACE - TagPrime & 0 & 0 & 0 & 0 & 0 & 0 & 0 & 0 & 0 \\
        % MEE - TagPrime & & & & & & & & & & & & & & & & \\
        DivED* & 0 & 0 & 16 & 25 & 32 & 1 & 0 & 7 & 10 \\
        Keyword* & 8 & 11 & 10 & 14 & 12 & 12 & 20 & 8 & 12 \\
        GPT-3.5-turbo* & 13 & 14 & 20 & 35 & 45 & 12 & 12 & 14 & 21 \\
        \midrule
        \multicolumn{10}{c}{\small \textsc{Trained on SPEED++ (Our Framework)}} \\
        \midrule
        TagPrime & 50 & 0 & 35 & 62 & \textbf{59} & \textbf{56} & 0 & 26 & \underline{36} \\
        TagPrime + XGear & 50 & 0 & 35 & 62 & \textbf{59} & \textbf{56} & 0 & 26 & 36 \\
        BERT-QA & 46 & 0 & 31 & 60 & 57 & 43 & 0 & 22 & 32 \\
        DyGIE++ & \textbf{51} & 0 & \textbf{36} & 62 & 58 & 54 & 0 & 21 & 35 \\
        OneIE & 50 & 0 & 35 & \textbf{63} & 58 & 55 & 0 & 23 & 36 \\
        \midrule
        % \midrule
        % \multicolumn{17}{c}{\small \textsc{Trained on SPEED++ (Our Framework) with CLaP}} \\
        % \midrule
        TagPrime + CLaP & 45 & \textbf{27} & \textbf{36} & 62 & \textbf{59} & \textbf{56} & \textbf{27} & \textbf{28} & \textbf{42} \\
        % TagPrime ED + XGear EAE & & & & & & & & & & & & & & & & \\
        % EEQA ED + EAE & & & & & & & & & & & & & & & & \\
        
        \bottomrule
    \end{tabular}
    \caption{Benchmarking EE models trained on \dataName{} for extracting event information in the cross-lingual cross-disease setting. The evaluation used is Trigger Classification (TC). Here, hi = Hindi, jp = Japanese, es = Spanish, and en = English. *Numbers are higher compared to others as evaluation is done using string matching.}
    \label{tab:tc-main-results}
\end{table}

\begin{table}[t]
    \centering
    \small
    \setlength{\tabcolsep}{1.9pt}
    \begin{tabular}{l|c|c|c|c|c|c|c|c|c}
        \toprule
        \multirow{3}{*}{\textbf{Model}} & \multicolumn{3}{c|}{\textbf{COVID}} & \multicolumn{1}{c|}{\textbf{MPox}} & \multicolumn{4}{c|}{\textbf{Zika + Dengue}} & \multicolumn{1}{c}{\textbf{Avg}} \\
        & \multicolumn{1}{c}{hi} & \multicolumn{1}{c}{jp} & \multicolumn{1}{c|}{es} & \multicolumn{1}{c|}{en} & \multicolumn{1}{c}{en} & \multicolumn{1}{c}{hi} & \multicolumn{1}{c}{jp} & \multicolumn{1}{c|}{es} \\ 
        \midrule
        % \multicolumn{10}{c}{\small \textsc{Trained on SPEED++ (Our Framework)}} \\
        % \midrule
        TagPrime & \textbf{55} & \textbf{21} & \textbf{51} & \textbf{58} & \textbf{67} & \textbf{61} & \textbf{25} & \textbf{50} & \textbf{49} \\
        XGear & 29 & 17 & 49 & 54 & 63 & 44 & 19 & 47 & 40 \\
        BERT-QA & \textbf{55} & 17 & 46 & 53 & 64 & 59 & 7 & 49 & \underline{44} \\
        % DyGIE++ & \textbf{51} & 0 & \textbf{36} & 62 & 58 & 54 & 0 & 21 & 35 \\
        % OneIE & 50 & 0 & 35 & \textbf{63} & 58 & 55 & 0 & 23 & 36 \\
        
        \bottomrule
    \end{tabular}
    \caption{Benchmarking EAE models trained on \dataName{} for extracting event information in the cross-lingual cross-disease setting. The evaluation used is Argument Classification (AC). Here, hi = Hindi, jp = Japanese, es = Spanish, and en = English.}
    \label{tab:eae-only-results}
\end{table}

\subsection{GPT-3}

We use the GPT-3.5 turbo model as the base GPT model.
% We experiment with ChatGPT \cite{chatgpt} for tuning our prompts that ensure output consistency.
We illustrate our final prompt template in Figure~\ref{fig:ee-gpt-prompt}.
It majorly comprises a task definition, ontology details, 1 example for each event type along with corresponding arguments, and the final test query.
We conducted a looser evaluation for GPT and only matched if the predicted trigger text matched the gold trigger text.

\begin{figure*}[t]
    \centering
    \includegraphics[width=0.9\textwidth]{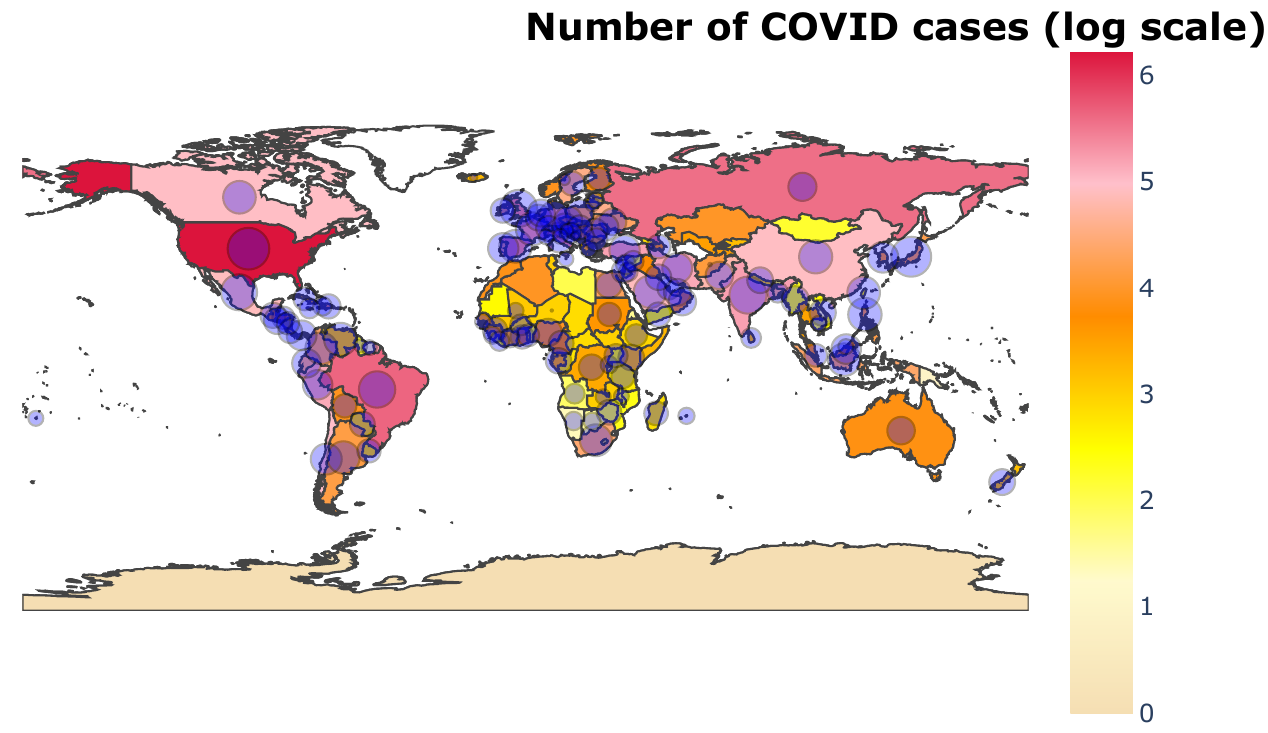}
    \caption{Geographical distribution for the whole world of the number of reported COVID-19 cases as of May 28, 2020. The blue dots indicate the events extracted by our model and the size indicates the number of epidemic events for the specific country (log scale).}
    \label{fig:world-map-plot}
\end{figure*}

\section{Additional Epidemic Event Extraction Experiments}
\label{sec:additional-eee-expts}

In addition to the performance evaluation of the event and argument classification of SPEED++, we benchmark its event \textit{trigger} classification (\textbf{TC}) performance using the same benchmarking settings.
More simply, TC is a stricter evaluation metric that computes the F1 score of the \textit{(trigger, event type)} pair.
% This task extracts the trigger word in a sentence that leads to the detection of an event of a certain type.
We present our results in Table~\ref{tab:tc-main-results}. We continue to observe the strongest overall performance by the \textbf{supervised baselines trained on our \dataName{} dataset with TagPrime}.
Models trained without CLaP perform poorly with a zero F1 score for Japanese.
This can be attributed to tokenization difference as Japanese is treated as a character-level language.
Since models are trained on English component of \dataName{}, it has a strong prior for trigger words to be a single token.
But for Japanese, each character is treated as a token and it possesses multi-token triggers.
Due to this mismatch, we note that the \dataName{}-trained models provide zero scores.
This is improved when additional training is done using augmented data from CLaP.
% Our models trained with unaugmented SPEED++ perform poorly on the trigger classification task when tested on the character-level tokenized Japanese data.
% likely due to the lack of models' realizations of the presence of trigger words comprising more than one token.

\paragraph{EAE Benchmarking}
We also benchmark our pure EAE models trained with \dataName{} and present our results in Table~\ref{tab:eae-only-results}.
These models are provided with the gold event annotations and required to predict all possible arguments corresponding to the gold events.
As evident from Table~\ref{tab:eae-only-results}, TagPrime model preforms the best across the different families of models across the different languages and diseases.

\section{Global Epidemic Prediction: Additional Details}
\label{sec:appendix-gep-additional-details}

To validate the breadth and coverage of our multilingual framework, we utilize it to detect COVID-19 pandemic-related events from social media.
Specifically, we focus on all tweets from a single day (chosen at random) - May 28, 2020.
We utilize Twitter's Language Identification for sorting the tweets into different languages, resulting in a total of 65 langauges.
Next, we map each tweet to a specific location and country, as explained below

\paragraph{Location mapping}
We utilize the user's location to map each tweet to a specific country.
For tweets without location specified, we pool them into a set of unspecified tweets.
We estimate the country distribution of tweets per language using the tweets with specified locations.
Utilizing this distribution, we extrapolate the locations to the unspecified tweets to approximate the actual location distribution of the tweets.
Mapping tweets from the 65 languages results in a country distribution over 117 countries worldwide.

\paragraph{Geographical plotting}
We consider the 117 countries mapped for these plots.
Majorly, we pan these countries out on a map and color them based on the number of COVID-19 cases reported\footnote{\url{https://www.worldometers.info/coronavirus}} until May 28, 2020.
Lighter shades indicate fewer cases, while darker shades indicate massive spread in those countries.
We utilize our framework to extract the number of events from the country-mapped tweets and plot them as transculent circles.
Bigger dots indicate more events extracted for the given country.
We show this geographically for the whole world in Figure~\ref{fig:world-map-plot} and just for Europe in Figure~\ref{fig:europe-map-plot}.

From the world map, we note how many of the red countries (United States of America, India, Brazil) have large dots associated with them - indicating more events found for countries where the spread of the disease was high.
Similarly, countries where the spread was lesser correspondingly have smaller dots - indicating lesser epidemic events found for these countries.
We observe a large cluster for Europe and thus plot it separately, shown in Figure~\ref{fig:europe-map-plot} and discussed in \S~\ref{sec:global-epidemic-prediction}.

% The countries are colored based on how many total cases were registered there as of May 28, 2020. Lighter shades of yellow indicate fewer cases, while darker shades of red indicate more cases. The size of a dot for any given country is based on how many total events were detected across the world for the main language spoken by the country; unfortunately, at present we do not have a solid way to differentiate between countries that have the same main language, so the number of events has been aggregated for all Spanish-speaking countries, all Arabic-speaking countries, all French-speaking countries, Portuguese-speaking countries, etc. This is why there are so many events detected across Latin America: all the countries where Spanish is spoken have been assigned the same set of events. There are two reasons why a country may not have an event count associated with it: either the number of events was too small to register on the plot, or the language associated with the country was not among the 26 languages we found events for.

\begin{table}[t]
    \centering
    \small
    \begin{tabular}{lp{4.5cm}r}
        \toprule
        \textbf{Rank} & \textbf{Clustered Argument} & \textbf{Count} \\
        \midrule
        \multicolumn{3}{c}{ \textsc{\redtext{English} \quad - \quad \bluetext{Monkeypox} \quad - \quad \browntext{Symptoms}}} \\
        \midrule
        1 & rash & 818 \\
        2 & sick & 746 \\
        3 & lesions & 637 \\
        4 & fever & 548 \\
        5 & side effect & 484 \\
        6 & itching & 441 \\
        7 & rashes & 412 \\
        8 & cough & 175 \\
        \midrule
        \multicolumn{3}{c}{ \textsc{\redtext{English} \quad - \quad \bluetext{Zika} \quad - \quad \browntext{Symptoms}}} \\
        \midrule
        1 & birth defects & 2.2K \\
        2 & brain damage & 1.1K \\
        3 & microcephaly & 990 \\
        4 & health problems & 723 \\
        5 & nerve disorder & 705 \\
        6 & congenital syndrome & 391 \\
        7 & nerve damage & 382 \\
        8 & damages placenta & 196 \\
        \midrule
        \multicolumn{3}{c}{ \textsc{\redtext{English} \quad - \quad \bluetext{Dengue} \quad - \quad \browntext{Symptoms}}} \\
        \midrule
        1 & fever & 4.8K \\
        2 & multiple organ failure & 1.1K \\
        3 & shock syndrome & 522 \\
        4 & symptoms & 326 \\
        5 & high fever & 292 \\
        6 & severe disease & 256 \\
        7 & disease & 250 \\
        8 & rashes & 200 \\
        \bottomrule
    \end{tabular}
    \caption{Aggregated information about symptoms for Monkeypox, Zika, and Dengue from English tweets using our \dataName{} framework.}
    \label{tab:info-assimilation-full-table-english-diseases}
\end{table}

\begin{table}[t]
    \centering
    \small
    \begin{tabular}{lp{4.5cm}r}
        \toprule
        \textbf{Rank} & \textbf{Clustered Argument} & \textbf{Count} \\
        \midrule
        \multicolumn{3}{c}{ \textsc{\redtext{English} \quad - \quad \bluetext{COVID-19} \quad - \quad \browntext{Symptoms}}} \\
        \midrule
        1 & can’t breathe & 8.8K \\
        2 & pneumonia & 6.7K \\
        3 & sick & 4.2K \\
        4 & hemorrhaging & 2.9K \\
        5 & prevents me from staying home & 2.7K \\
        6 & cough & 2.1K \\
        7 & symptoms & 1.9K \\
        8 & critically ill & 1.2K \\
        \midrule
        \multicolumn{3}{c}{ \textsc{\redtext{English} \quad - \quad \bluetext{COVID-19} \quad - \quad \browntext{Cure Measures}}} \\
        \midrule
        1 & hydroxychloroquine & 3.7K \\
        2 & remdesivir & 2.5K \\
        3 & drug & 2.1K \\
        4 & treatment & 1.7K \\
        5 & hcq & 1.3K \\
        6 & vaccine & 485 \\
        7 & zinc & 448 \\
        8 & lockdown & 425 \\
        \midrule
        \multicolumn{3}{c}{ \textsc{\redtext{English} \quad - \quad \bluetext{COVID-19} \quad - \quad \browntext{Control Measures}}} \\
        \midrule
        1 & lockdown & 187K \\
        2 & quarantine & 56K \\
        3 & social distancing & 38K \\
        4 & deny entry & 28K \\
        5 & response & 21K \\
        6 & title 32 orders & 15K \\
        7 & masks & 15K \\
        8 & executive order & 10K \\
        \bottomrule
    \end{tabular}
    \caption{Aggregated information about various arguments for COVID-19 from English tweets using our \dataName{} framework.}
    \label{tab:info-assimilation-full-table-english-covid}
\end{table}

\section{Epidemic Information Aggregation: Additional Details}
\label{sec:appendix-eia-additional-details}

In \S~\ref{sec:epidemic-info-assimilation}, we discussed how we utilize the EAE capability of our trained TagPrime model for creating an information aggregation bulletin.
Here we specify more details about this process.
First, we utilize our EE framework to extract all possible arguments for the event-specific roles.
Since many similar arguments can be extracted, we merge them together by clustering.
To this end, we project the arguments into a higher-dimensional embedding space using a Sentence Transformer\footnote{\url{https://huggingface.co/sentence-transformers/distiluse-base-multilingual-cased-v2}} \cite{reimers-2019-sentence-bert} encoding model.
Next, we utilize a hierarchical agglomerative clustering (HAC) algorithm to merge similar arguments.
We implement the clustering using sklearn\footnote{\url{https://scikit-learn.org/stable/modules/generated/sklearn.cluster.AgglomerativeClustering.html}} utilizing euclidean distance as the distance metric and a threshold of 1 as the stopping criteria.
After generating the clusters, we rank the clusters by the occurrence count of all arguments in the cluster and label them based on the most frequent argument.

We report the top-ranked clustered arguments for several event roles for COVID-19 from English tweets in Table~\ref{tab:info-assimilation-full-table-english-covid}.
We report similar tables for different diseases from English tweets in Table~\ref{tab:info-assimilation-full-table-english-diseases} and COVID-19 from multilingual tweets in Table~\ref{tab:info-assimilation-full-table-multi-covid}.
Despite some irrelevant extractions owing to model inaccuracies, most of these top clustered arguments are relevant and reflect the language and disease-specific properties quite accurately.

\begin{table}[t]
    \centering
    \small
    \begin{tabular}{lp{2.5cm}p{2cm}r}
        \toprule
        \textbf{Rank} & \textbf{Argument} & \textbf{Translation} & \textbf{Count} \\
        \midrule
        \multicolumn{4}{c}{ \textsc{\redtext{Hindi} \quad - \quad \bluetext{COVID-19} \quad - \quad \browntext{Cure Measures}}} \\
        \midrule
        1 & \hi{ilAj} & treatment & 1.1K \\
        2 & \hi{hom aAisol\?fn} & home isolation & 1K \\
        3 & \hi{yog} & yoga & 636 \\
        4 & \hi{-vA-Ly lAB} & recover & 500 \\
        5 & \hi{gO\8{m}/ k\? \7{k}\3A5w\4} & cow urine rinse & 448 \\
        6 & \hi{aApk\? aAfFvA\0d} & your blessings & 240 \\
        7 & \hi{EX-cAj\0} & discharge & 126 \\
        8 & \hi{dvAao\2} & medicines & 120 \\
        \midrule
        \multicolumn{4}{c}{ \textsc{\redtext{Spanish} \quad - \quad \bluetext{COVID-19} \quad - \quad \browntext{Cure Measures}}} \\
        \midrule
        1 & hidroxicloroquina & hydroxychloroquine & 583 \\
        2 & leche materna & breastmilk & 427 \\
        3 & medicamentos & medicines & 252 \\
        4 & tratamientos & treatments & 226 \\
        5 & red integrada covid & covid integrated network & 214 \\
        6 & ivermectina & ivermectin & 157 \\
        7 & remdesivir & remdesivir & 152 \\
        8 & transplante & transplant & 132 \\
        \bottomrule
    \end{tabular}
    \caption{Aggregated information about various arguments for Hindi and Spanish for COVID-19 using our \dataName{} framework.}
    \label{tab:info-assimilation-full-table-multi-covid}
\end{table}

\paragraph{Example tweets}
We also provide qualitative example tweets mentioning some of these arguments to prove the efficacy of our EAE framework.
Table~\ref{tab:info-assimilation-examples-eng-covid} presents various English tweets for COVID-19 related mentions.
Table~\ref{tab:info-assimilation-examples-eng-diseases} presents various English tweets for other diseases of Monkeypox, Zika, and Dengue with their mentions.
Table~\ref{tab:info-assimilation-examples-multi-covid} presents various Hindi and Spanish tweets for COVID-19-related mentions.
Through this table, we see the diverse set of tweets and how our framework can extract these arguments across them.

\begin{table}[t]
    \centering
    \small
    \begin{tabular}{p{7.2cm}}
        \toprule
        \multicolumn{1}{c}{ \textbf{Tweet} } \\
        \midrule
        \multicolumn{1}{c}{ \textsc{\redtext{English} \quad - \quad \bluetext{COVID-19} \quad - \quad \browntext{Symptoms}} } \\
        \midrule
        My mum has \redtext{pneumonia} and it might be because of corona, praying for her man \\ \hline
        Autopsies of African Americans who died of \#(COVID) in New Orleans reveal \redtext{hemorrhaging} ... \\ \hline
        Apply for a test if you have symptoms of \#(coronavirus): a high temperature, a new continuous \redtext{cough}, loss or change to your sense of smell or taste \\
        \midrule
        \multicolumn{1}{c}{ \textsc{\redtext{English} \quad - \quad \bluetext{COVID-19} \quad - \quad \browntext{Cure Measures}} } \\
        \midrule
        Trump reveals he's taking \redtext{hydroxychloroquine} in effort to prevent and cure coronavirus symptoms \\ \hline
        In this new Covid-19 audio interview, editors discuss newly published studies of \redtext{remdesivir} that highlight its potential and its problems \\ \hline
        Hydroxychloroquine combined with \redtext{zinc} has shown effective in treating covid-19 \\
        \midrule
        \multicolumn{1}{c}{ \textsc{\redtext{English} \quad - \quad \bluetext{COVID-19} \quad - \quad \browntext{Control Measures}} } \\
        \midrule
        We should not return to school, we should not undo any other aspect of the \redtext{lockdown} until the test, trace and isolation policy is fully in place \\ \hline
        Boris Johnson says from Monday, up to six people will be allowed to meet outside subject to \redtext{social distancing} rules in England \\ \hline
        Masks work. Everyone has to wear a \redtext{mask} when in any business ... \\
        \bottomrule
    \end{tabular}
    \caption{Illustration of actual tweets in English mentioning various symptoms, cure measures, and control measures related to COVID-19. The terms extracted by our system are highlighted in \redtext{red}.}
    \label{tab:info-assimilation-examples-eng-covid}
\end{table}

\begin{table}[t]
    \centering
    \small
    \begin{tabular}{p{7.2cm}}
        \toprule
        \multicolumn{1}{c}{ \textbf{Tweet} } \\
        \midrule
        \multicolumn{1}{c}{ \textsc{\redtext{English} \quad - \quad \bluetext{Monkeypox} \quad - \quad \browntext{Symptoms}} } \\
        \midrule
        I have a pretty mild \redtext{rash} on my stomach. A little bit of itchy. The extremely optimistic part of my brain is like \"What if it's monkey pox? \\ \hline
        Anyone can get \#(monkey pox) through close skin-to-skin contact ... Healthcare providers must be vigilant and test any patient with a suspicious \redtext{lesion} or sore. \\ \hline
        Its inappropriate to say but the amount of \redtext{itching} Ive done from the bites makes me nervous people are gonna think Ive got monkey pox or some shit \\
        \midrule
        \multicolumn{1}{c}{ \textsc{\redtext{English} \quad - \quad \bluetext{Zika} \quad - \quad \browntext{Symptoms}} } \\
        \midrule
        More \redtext{birth defects} seen in (url) areas where Zika was present (url) \\ \hline
        Zika \redtext{brain damage} may go undetected in pregnancy \\ \hline
        study sheds light on how Zika causes \redtext{nerve disorder} (url) \\
        \midrule
        \multicolumn{1}{c}{ \textsc{\redtext{English} \quad - \quad \bluetext{Dengue} \quad - \quad \browntext{Symptoms}} } \\
        \midrule
        In the evening the \redtext{fever} is skyrocketing \& the joint pain is born-breaking \& nauseating, vomiting is constant \\ \hline
        Dengvaxia = yellow fever vaccine + live attenuated dengue virus. \redtext{Multiple organ failure} was already established as its key symptom \\ \hline
        TMI but I’ve had \redtext{rashes} on my arms and legs for a couple of days now. Tried to tell Scott I have dengue fever but ... \\
        \bottomrule
    \end{tabular}
    \caption{Illustration of actual tweets in English mentioning various symptoms related to Monkeypox, Zika, and Dengue. The terms extracted by our system are highlighted in \redtext{red}.}
    \label{tab:info-assimilation-examples-eng-diseases}
\end{table}

\begin{table*}[h]
\centering
\small
\begin{tabular}{p{2.5cm}|p{6cm}|p{6cm}}
\toprule
\multicolumn{3}{c}{Arguments of INFECT event}                                                                                                                                             \\ \midrule
\textbf{Argument Name} & \textbf{Argument Definition} & \textbf{Example} \\ \midrule
Infected         & The individual(s) being infected & 300 \underline{people} tested positive.                       \\ \midrule
Disease            & The disease or virus that invaded the host & I tested positive for \underline{COVID}.                                                                                   \\ \midrule
Place              & The place that the individual(s) are infected & 5 students at \underline{school} are infected.                                                \\ \midrule
Time               & The time that the individual(s) are infected & 300 people tested positive on \underline{May 15}.                                \\ \midrule
Value              & The number of people being infected & \underline{Some} people are infected.                                                 \\ \midrule
Information-source & The source that is providing this information regarding to infection & According to \underline{CDC}, if you have COVID...                               \\ \bottomrule
\end{tabular}
\caption{Complete definition and examples of arguments of INFECT event}
\label{tab:infect-ontology}
\end{table*}

\begin{table*}[h]
\centering
\small
\begin{tabular}{p{2.5cm}|p{6cm}|p{6cm}}
\toprule
\multicolumn{3}{c}{Arguments of SPREAD event}                                                                                                                                             \\ \midrule
\textbf{Argument Name} & \textbf{Argument Definition} & \textbf{Example} \\ \midrule
Population         & The population among which the disease spreads & 16000 \underline{Americans} are infected.                       \\ \midrule
Disease            & The disease/virus/pandemic that is prevailing & \underline{Monkey-box} is spreading ...                                                                                   \\ \midrule
Place              & The place at which the disease is spreading & the flu prevails in \underline{the U.S.}.                                                \\ \midrule
Time               & The time during which the disease is spreading & the flu prevails in the U.S. in \underline{winter}.                                \\ \midrule
Value              & The number of people being infected & \underline{16000} Americans are infected.                                                 \\ \midrule
Information-source & The source that is providing this information regarding to the transmission of the disease & \underline{My mom} says COVID is spreading again.                                 \\ \midrule
Trend              & The possible change of a transmission of a disease with respect to past status & COVID \underline{spreads faster than we've expected}.\\ \bottomrule
\end{tabular}
\caption{Complete definition and examples of arguments of SPREAD event}
\label{tab:spread-ontology}
\end{table*}

\begin{table*}[h]
\centering
\small
\begin{tabular}{p{2.5cm}|p{6cm}|p{6cm}}
\toprule
\multicolumn{3}{c}{Arguments of SYMPTOM event}                                                                                                                                             \\ \midrule
\textbf{Argument Name} & \textbf{Argument Definition} & \textbf{Example} \\ \midrule
Person         & The individual(s) displaying symptoms &  \underline{I'm} coughing now.                     \\ \midrule
Symptom & The concrete symptom(s) that are displayed & You may have \underline{severe fever} and \underline{stomach-ache}.\\ \midrule
Disease            & The disease(s)/virus that are potentially causing the symptoms & If you cough, that's probably \underline{COVID}.                                                                                   \\ \midrule
Place              & The place at which the symptom(s) are displayed & Students are showing illness at \underline{school}.                                                \\ \midrule
Time               & The time during which the symptom(s) are displayed & I feel sick \underline{yesterday}.                                \\ \midrule
Duration              & The time interval that the symptom(s) last & My fever lasts \underline{three days}.                   \\ \midrule
Information-source & The source that is providing this information regarding to symptoms of the disease & \underline{He} said half of his class were ill.                                 \\  \bottomrule
\end{tabular}
\caption{Complete definition and examples of arguments of SYMPTOM event}
\label{tab:symptom-ontology}
\end{table*}

\begin{table*}[h]
\centering
\small
\begin{tabular}{p{2.5cm}|p{6cm}|p{6cm}}
\toprule
\multicolumn{3}{c}{Arguments of PREVENT event}                                                                                                                                             \\ \midrule
\textbf{Argument Name} & \textbf{Argument Definition} & \textbf{Example} \\ \midrule
Agent         & The individual(s) attempting to avoid infections  &  \underline{You} should wear a mask to protect yourself and others.   \\ \midrule
Disease            & The disease/virus/illness being defensed against & prevent \underline{COVID} infection.                                                                                   \\ \midrule
Means              & actions/means that may prevent infection & You should \underline{wear a mask} to protect yourself and others.                                                \\ \midrule
Information-source & The source that is providing this information regarding to the prevention of this disease & \underline{CDC} proves masks can efficiently blocks virus.                                 \\ \midrule
Target & The individual(s)/population to which the agent attempts to prevent the disease transmission & You should wear a mask to protect \underline{yourself} and \underline{others}.\\ \midrule
Effectiveness & How effective is the means against the disease & CDC proves masks can \underline{efficiently blocks virus}. \\
\bottomrule
\end{tabular}
\caption{Complete definition and examples of arguments of PREVENT event}
\label{tab:prevent-ontology}
\end{table*}

\begin{table*}[h]
\centering
\small
\begin{tabular}{p{2.5cm}|p{6cm}|p{6cm}}
\toprule
\multicolumn{3}{c}{Arguments of CONTROL event}                                                                                                                                             \\ \midrule
\textbf{Argument Name} & \textbf{Argument Definition} & \textbf{Example} \\ \midrule
Authority         & The authority implementing/advocating the control of a pandemic  &  To impede COVID transmission, \underline{Chinese government} required quarantine upon arrival.   \\ \midrule
Disease            & The intruding disease/virus/pandemic being defensed against & To impede \underline{COVID} transmission... \\ \midrule
Means              & the enacted/advocated policies/actions that may control the pandemic & To impede COVID transmission, Chinese government required \underline{quarantine} upon arrival.                                                \\ \midrule
Information-source & The source that is providing this information regarding to the control of this disease & \underline{CNN} reports massive pandemic lockdowns in China.                                 \\ \midrule
Place & The individual(s)/population to which the agent attempts to prevent the disease transmission & \underline{NY}\underline{ wil}l enforces mask policy from June.\\ \midrule
Time & The individual(s)/population to which the agent attempts to prevent the disease transmission & NY will enforces mask policy from \underline{June}.\\ \midrule
Effectiveness & How effective is the means against the disease & The \underline{infection rate does not decrease} since the enforcement of mask policy. \\ \midrule
Subject &The individual(s)/population encouraged/ordered to implement the control measures&Due to the pandemic, \underline{students} are required to wear masks in class. \\
\bottomrule
\end{tabular}
\caption{Complete definition and examples of arguments of CONTROL event}
\label{tab:control-ontology}
\end{table*}

\begin{table*}[h]
\centering
\small
\begin{tabular}{p{2.5cm}|p{6cm}|p{6cm}}
\toprule
\multicolumn{3}{c}{Arguments of CURE event}                                                                                                                                             \\ \midrule
\textbf{Argument Name} & \textbf{Argument Definition} & \textbf{Example} \\ \midrule
Cured         & The individuals(s) recovered/receiving the treatments  &  \underline{My grandma} recovered from COVID yesterday.   \\ \midrule
Disease            & The disease/illness that the patients get rid of & My grandma recovered from \underline{COVID} yesterday. \\ \midrule
Means              & The therapy that (potentially) treat the disease & Just \underline{get rest} and your fever will go away.                                                \\ \midrule
Information-source & The source that is providing this information regarding to the cure/recovery of this disease & \underline{CNN} reports that XX company claimed to developed COVID treatment.    \\ \midrule
Place & The place at which the recovery takes place & In \underline{the U.S.}, 15670 people recovered and 16000 died of COVID.\\ \midrule
Time & The time at which the recovery takes place & By \underline{May 15}, 15670 Americans recovered and 16000 died of COVID.\\ \midrule
Effectiveness & How effective is the means against the disease  & The new COVID treatment is \underline{not fully effective}. \\ \midrule
Value &The number of people being cured & By May 15, \underline{15670} Americans recovered and 16000 died of COVID.\\ \midrule
Facility &The individual(s)/organization(s) utilizing/inventing certain means to facilitate recoveries & CNN reports that \underline{XX company} claimed to developed COVID treatment. \\ \midrule
Duration &The time interval that the treatment takes & I received the treatment for \underline{two month} before full recovery. \\ 
\bottomrule
\end{tabular}
\caption{Complete definition and examples of arguments of CURE event}
\label{tab:cure-ontology}
\end{table*}

\begin{table*}[h]
\centering
\small
\begin{tabular}{p{2.5cm}|p{6cm}|p{6cm}}
\toprule
\multicolumn{3}{c}{Arguments of DEATH event}                                                                                                                                             \\ \midrule
\textbf{Argument Name} & \textbf{Argument Definition} & \textbf{Example} \\ \midrule
Dead         & The individuals(s) who die of infectious disease &  By March, 500 \underline{people} died of COVID in CA.   \\ \midrule
Disease            & The disease/virus/pandemic that (potentially) causes the death & By March, 500 people died of the \underline{virus} in CA. \\ \midrule
Information-source & The source that is providing this information regarding to fatality of this disease & \underline{Daily news}: new death of COVID ... \\ \midrule
Place & The place at which the death takes place & By March, 500 people died of COVID in \underline{CA}.\\ \midrule
Time & The time at which the death takes place & By \underline{March}, 500 people died of COVID in CA.\\ \midrule
Value & The number of death due to infectious disease & By March, \underline{500} people died of COVID in CA.\\ \midrule
Trend &The possible change of death counts caused by disease infection compared to the statistics from the past. & The COVID death toll is \underline{still increasing...} \\ 
\bottomrule
\end{tabular}
\caption{Complete definition and examples of arguments of DEATH event}
\label{tab:death-ontology}
\end{table*}

\begin{table*}[t]
    \centering
    \small
    \begin{tabular}{l|p{3.1cm}|p{3.1cm}|p{3.1cm}|p{3.1cm}}
        \toprule
        \textbf{Event} & \textbf{English} & \textbf{Hindi} & \textbf{Spanish} & \textbf{Japanese} \\
        \midrule
        \multirow{4}{*}{Infect} & I \redtext{caught} the virus earlier today & \hi{m\4{\qva} aAj \7{s}bh vAyrs s\? \redtext{bFmAr} ho gyA \8{h}\2} & \redtext{Contraje} el virus temprano hoy & \ja{今日ウイルスに\redtext{感染した}} \\ \cline{2-5}
        & My brother tested \redtext{positive} for COVID-19 & {\dn m\?r\? BAI kA} COVID-19 {\dn V\?-V \redtext{pA\<EjEVv} aAyA} & Mi hermano dio \redtext{positivo} por COVID-19. & \ja{私の兄はCOVID-19\redtext{陽性}でした} \\ \hline
        \multirow{8}{*}{Spread} & The COVID-19 \redtext{outbreak} put WHO in alert that the pandemic may develop into global scale & COVID-19 {\dn k\? \redtext{\3FEwkop} n\?} WHO {\dn ko stk\0 kr EdyA h\4 Ek yh mhAmArF v\4E\398wk -tr pr EvkEst ho sktF} & El \redtext{brote} Covid-19 alertó al OMS que la pandemia puede alcanzar en escala global & \ja{COVID-19の\redtext{発生}により、WHOはパンデミックが世界的な規模に発展する可能性に警戒を強めている} \\ \cline{2-5}
        & A new flu is \redtext{sweeping} across Los Angeles & \hi{lA\<s e\2EjSs m\?{\qva} ek nyA \325w\8{l} \redtext{P\4l} rhA h\4} & Una nueva gripe está \redtext{propagando} a traves de Los Ángeles & \ja{ロサンゼルスで新型インフルエンザが\redtext{流行}} \\ \hline
        \multirow{6}{*}{Symptom} & Many of my friends have a \redtext{cold} & \hi{m\?r\? k\4i do-to{\qva} ko \redtext{sdF{\qvb}} h\4} & Muchos de mis amigos tienen un \redtext{resfriado} & \ja{私の多くの友人が風邪を\redtext{ひいた}} \\ \cline{2-5}
        & I became incredibly \redtext{ill} after catching the virus & {\dn vAyrs kF cp\?V m\?{\qva} aAn\? k\? bAd m\4{\qva} aEv\398wsnFy !p s\? \redtext{bFmAr} ho gyA}  & Me \redtext{enfermé} increíblemente después de contagiarme de el virus & \ja{ウイルスに感染した後、私はすごく体調を\redtext{崩した}} \\ \hline
        \multirow{6}{*}{Prevent} & Medical experts encourage young kids to \redtext{wash} their hands & \hi{EcEk(sA Evf\?q\3E2w CoV\? b\3CEwo{\qva} ko hAT \redtext{Don\?} k\? Ele \3FEwo(sAEht krt\? h\4{\qva}} & Los expertos médicos alientan a los niños a \redtext{lavarse} las manos & \ja{医療専門家が幼児に\redtext{手洗い}を奨励} \\ \cline{2-5}
        & Wear a mask to \redtext{protect} your family from the disease & {\dn apn\? pErvAr ko \redtext{bFmArF} s\? bcAn\? k\? Ele mA-k phn\?{\qva}} & Use una máscara para \redtext{proteger} a su familia de la enfermedad & \ja{家族を疾病から\redtext{守る}ためにマスクを着用すること} \\ \hline
        \multirow{8}{*}{Control} & The WHO has published new \redtext{guidelines} in response to the rising cases of COVID-19 & WHO {\dn n\?} COVID-19 {\dn k\? bxt\? mAmlo{\qva} k\? jvAb m\?{\qva} ne \redtext{EdfAEnd\?{\qvb}f} \3FEwkAEft Eke h\4{\qva}} & La OMS ha publicado nuevas \redtext{pautas} en respuesta a los casos crecientes de Covid-19 & \ja{WHOはCOVID-19の感染者増加を受けて新しい\redtext{ガイドライン}を発表した。} \\ \cline{2-5}
        & Government officials have \redtext{imposed} a lockdown on certain districts & \hi{srkArF aEDkAEryo{\qva} n\? \7{k}C Ejlo{\qva} m\?{\qva} lA\<kXAun \redtext{lgA} EdyA h\4} & Los funcionarios gubernamentales han \redtext{impuesto} un aislamiento a ciertos distritos & \ja{政府当局が特定の地区にロックダウンを\redtext{課した}} \\ \hline
        \multirow{5}{*}{Cure} & There is no magic \redtext{cure} for the pandemic & \hi{aBF tk koEvX kA koI \3FEwBAvF \redtext{ilAj} nhF{\qva}} & No existe una \redtext{cura} mágica para la pandemia & \ja{パンデミックに\redtext{特効薬}はない} \\ \cline{2-5}
        & Unfortunately doctors were unable to \redtext{save} him from the pandemic & \hi{\7{d}BA\0`y s\? XA\<?Vr us\? mhAmArF s\? \redtext{bcAn\?} m\?{\qva} asmT\0 T\?} & Desfortunadamente, los médicos no pudieron \redtext{salvarlo} de la pandemia & \ja{残念ながら、医師たちは彼をパンデミックから\redtext{救う}ことはできなかった。} \\ \hline
        \multirow{8}{*}{Death} & 700 people \redtext{killed} by COVID & \hi{koEvX s\? \rn{700} logo{\qva} kF \redtext{mOt}} & 700 personas \redtext{matadas} por Covid & \ja{COVIDによる\redtext{死亡率}：７００人} \\ \cline{2-5}
        & The \redtext{mortality} rate of the pandemic has decreased as experts figure out how to treat it & {\dn mhAmArF kF \redtext{\9{m}(\7{y} dr} m\?{\qva} kmF aAI h\4 \3C8wo{\qva}Ek Evf\?q\3E2w yh ptA lgA rh\? h\4{\qva} Ek iskA ilAj k\4s\? EkyA jAe} & La tasa de \redtext{mortalidad} de la pandemia ha disminuido a medida porque los expertos están descubriendo cómo tratarla & \ja{パンデミックの\redtext{死亡率}は、専門家による治療法の解明のために減少している。} \\
        \bottomrule
    \end{tabular}
    \caption{Sample translated seed tweets for the different event
types in our ontology for the different languages. Triggers are highlighted in \redtext{red}.}
    \label{tab:seed-tweets}
\end{table*}

\end{document}